\documentclass[10pt]{article}

\usepackage[margin=1in]{geometry}
\usepackage{amsmath,amssymb,mathtools}
\usepackage{booktabs}
\usepackage{array}
\usepackage{enumitem}
\usepackage{microtype}
\usepackage[hidelinks]{hyperref}
\usepackage{xcolor}
\usepackage{graphicx}
\usepackage{float}
\usepackage{placeins}

\title{\textbf{One Mistake, Three Layers: Persistence Rules Under Depth Correlation}}
\author{Maryam Rahimimovassagh \and  Michael Brown \and Sanjeda Sara Jennifer \and
        Niloofar Yousefi \\
        University of Central Florida}
\date{\today}
\newcommand{\ConSf}{-0.069}
\newcommand{\ConSfCI}{[-0.091, -0.048]}
\newcommand{\ConSp}{+0.110}
\newcommand{\ConSpCI}{[+0.091, +0.129]}
\newcommand{\ConSfDir}{negatively}
\newcommand{\ConSpDir}{positively}
\newcommand{\ConRatio}{1.6}
\newcommand{\ConNFit}{534}
\newcommand{\ConZeroPct}{49}
\newcommand{\ConMedTest}{1}
\newcommand{\ConMedAll}{51}

\newcommand{\ConCeiling}{21}
\newcommand{\ConInterNote}{The interaction does not separate from zero, so each slope is read at the other axis mean}

\newcommand{\CovAdj}{0.221}
\newcommand{\CovAdjCI}{[-0.279, -0.160]}
\newcommand{\CovUnadj}{-0.245}
\newcommand{\CovP}{0.004}
\newcommand{\CovFNPlace}{third most negative}
\newcommand{\CovMostNeg}{\(FF\) and \(NN\)}
\newcommand{\CovLeastNeg}{\(FN\) and \(NF\)}
\newcommand{\CovSplitEst}{-0.256}
\newcommand{\CovSplitCI}{[-0.429, +0.242]}
\newcommand{\CovLowestBase}{\(FN\)}

\newcommand{\MsaN}{534}
\newcommand{\MsaCeiling}{8,192}
\newcommand{\MsaNCeiling}{247}
\newcommand{\MsaNUncapped}{287}
\newcommand{\MsaRhoAll}{0.048}
\newcommand{\MsaRhoUncapped}{0.068}
\newcommand{\MsaRhoOthers}{0.038 and 0.063 for \texttt{both}; 0.046 and 0.086 for \texttt{zpool}}
\newcommand{\MsaFnDepth}{7,692}
\newcommand{\MsaNfDepth}{7,966}
\newcommand{\MsaFnCeil}{0.44}
\newcommand{\MsaNfCeil}{0.48}
\newcommand{\MsaAuprcRatio}{2.6}

\newcommand{\MsaShallowRegime}{\(NN\)}
\newcommand{\MsaShallowDepth}{5,156}
\newcommand{\MsaShallowAuprc}{0.476}

\newcommand{\HolmFamily}{18}
\newcommand{\HolmAlpha}{0.05}
\newcommand{\HolmNBoot}{2,000}

\newcommand{\HolmNFocus}{9}
\newcommand{\HolmNFocusSurvive}{9}

\newcommand{\HolmNOther}{9}
\newcommand{\HolmNOtherSurvive}{0}
\newcommand{\HolmNAtFloor}{7}
\newcommand{\HolmFloor}{0.0005}
\newcommand{\HolmWorstFocus}{0.015}

\begin{document}

\maketitle

\begin{abstract}
Probing frozen representations layer by layer is now standard practice,
but reporting when a signal first becomes decodable requires a
rule for reading a depth curve, and the obvious rule is wrong. On
synthetic curves with known ground truth, requiring a probe score to
exceed a null-calibrated threshold for three consecutive stages reports
an onset nineteen stages early under transformer-like depth
correlation, with every resample returning an onset. The failure follows from correlation between neighboring layers, a hazard well known in neuroimaging
but not, to our knowledge, checked for depth curves, and it is invisible to per-layer significance testing.

We give an estimator whose error rate is calibrated rather than assumed. Null trajectories are
constructed by permuting labels within each evaluation unit and holding
each permutation fixed across depth, which preserves the correlation
structure the rule must survive; thresholds are finite-sample order
statistics computed per layer; and the persistence requirement is
calibrated rather than chosen. We measure the resulting false-positive
rate on the study's own null using a leave-one-out construction, and
report an accumulation-depth statistic for regimes whose onset does not
resolve.

We apply the estimator to pocket information in a frozen co-folding
trunk, stratifying 1{,}820 post-cutoff targets by protein-family
familiarity and pocket familiarity. On this study's own null the
five-stage rule fires on 4 to 7 per cent of null trajectories against a
nominal 5 per cent, and the three-stage rule on 12 to 17: the synthetic
failure reproduces on real depth trajectories. Pocket membership proves
decodable above the calibrated threshold at every one of the 48 blocks
in all four regimes, so the onset itself cannot discriminate between
them; the pre-declared accumulation depth does, and is latest for
familiar families with structurally novel pockets in every feature set.
That regime is also the weakest at its peak, by a margin that survives
controls for protein length, the system-level pocket measure and
alignment depth; it is
the lowest regime at every pocket-familiarity threshold tested, though
the size of its deficit varies with that choice.

Separately, we report that a widely used public co-folding checkpoint
loads against its own published code with 128 trunk parameters
missing; those would be randomly initialized, and every representation
extracted from that pairing would be partly noise while appearing
entirely normal.
\end{abstract}
\clearpage

\section{Introduction}

Identifying ligand-binding pockets is a foundational step in
structure-based drug design, and co-folding models have changed how it
is done. Rather than scoring surface geometry, recent work reads
binding sites out of the representations these models build
internally: light per-residue heads on a frozen co-folding trunk
outperform geometric and graph-based detectors by a wide margin, and
the pocket signal lives in the pair representation rather than the
single-residue track \cite{zpocket}. That work also established the
axis along which such predictors generalise. Structural similarity
between a test pocket and the pockets a model has seen predicts
accuracy far better than sequence similarity between the proteins, and
accuracy collapses on pockets with no close structural precedent. We
take both results as given rather than as our question.

Reading one final representation leaves two things open. A co-folding
trunk refines its representation over dozens of blocks, and a
predictor trained on the last one cannot say where pocket information
becomes usable: whether it is present from the start and weakly
decodable, or assembled late. Those are different objects with
different implications for where one would intervene. And the collapse
on novel pockets has not been decomposed. A pocket can be novel
because its protein family is unfamiliar, because the pocket geometry
is unfamiliar, or both, and these need not be the same failure.
Separating them requires crossing the two axes rather than sweeping
one.

\begin{figure*}[t]
\centering
\includegraphics[width=\textwidth]{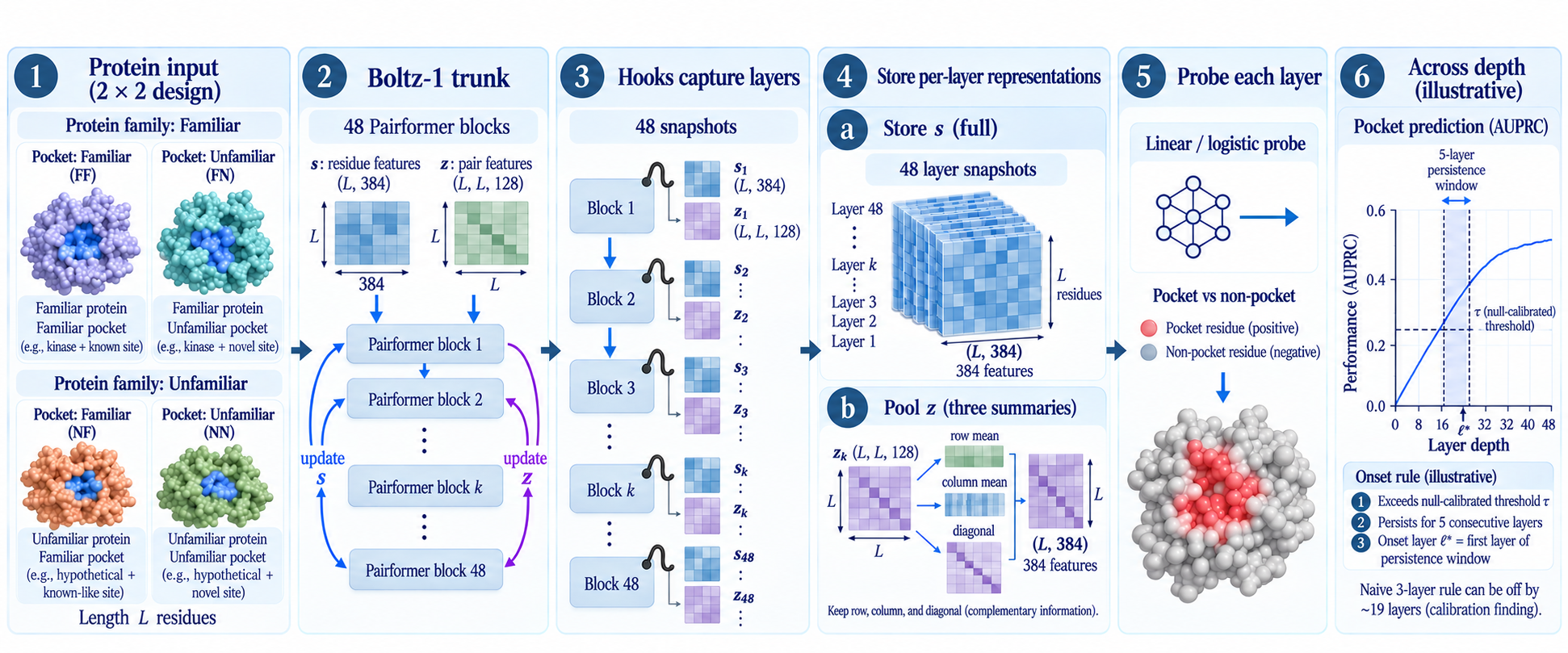}
\caption{%
Measuring when pocket information becomes decodable across trunk depth.
\textbf{(1)} Targets are assigned to four regimes by crossing
protein-family with pocket familiarity, both measured against pre-cutoff
systems. \textbf{(2)} The frozen Boltz-1 trunk refines
\(s\in\mathbb{R}^{L\times384}\) and \(z\in\mathbb{R}^{L\times L\times128}\)
through 48 Pairformer blocks; only the final recycle is retained.
\textbf{(3)} A forward hook on each block captures its \(s\) and \(z\).
\textbf{(4)} \(s\) is kept at full width; \(z\) is reduced to its row
mean, column mean and self-pair diagonal, the diagonal being
unrecoverable from either marginal. \textbf{(5)} A probe is fitted
per layer, scored by AUPRC since positives are about one residue in
fifteen. \textbf{(6)} The onset \(\ell^*\) is the first layer of the
earliest five-layer window above a permutation-calibrated threshold
\(\tau\). \emph{Panel~6 is illustrative and contains no measured data.}
Five layers rather than three: Section~\ref{sec:onset_validation} shows a
three-layer rule reporting onset nineteen stages early under
transformer-like depth correlation.%
}
\label{fig:pipeline}
\end{figure*}
Asking the first question safely is a prerequisite rather than a
detail. The obvious way to report when a signal emerges, the first
depth at which a probe clears a null threshold and stays above it for
a few consecutive layers, is badly wrong on transformer
representations, because adjacent blocks are residual updates of one
another and their probe scores are strongly correlated. On synthetic
curves with a known onset at stage 20, a three-consecutive-layer rule
reports emergence at stage 1. Every bootstrap resample returns an
onset, and the resulting interval spans almost the entire stack, so
the failure is not announced by the uncertainty estimate: it yields a
confident-looking number whose interval is too wide to contradict it.
We therefore construct a null-calibrated onset rule whose permutation
nulls are held fixed across depth and whose persistence requirement is
calibrated against that correlation, and we fix it before touching any
real representation.

We apply this to pocket information in a frozen co-folding trunk,
probing all 48 Pairformer blocks of 1{,}820 post-cutoff targets and
crossing protein-family familiarity with pocket familiarity to give
four regimes, \(FF\), \(FN\), \(NF\) and \(NN\); 534 of those targets
are held out for evaluation. We contribute three things. First, a
depth-resolved account: pocket membership is linearly decodable from
the first block in every regime, so the regimes differ in signal
strength and where it accumulates rather than in when it emerges.
Second, a sharper reading of what makes a pocket hard. The single
weakest cell is \(FN\), a familiar family with a structurally novel
pocket, peaking at 0.270 against 0.476 for \(NN\), the cell novel on
both axes, each read at its own peak layer; against its
confound-matched counterpart \(NF\) it trails by a factor of 2.6 at
the common evaluation layer. This specialises the known
structural-over-sequence axis into one counterintuitive cell, and it
is exactly the regime of finding new or allosteric pockets on
well-studied targets. Third, the onset rule itself, with the
demonstration that the naive alternative fails by nineteen stages.
What is new here is where across depth the information is usable,
which kind of novelty is hardest, and how to ask the depth question
without manufacturing a false onset.

\section{Related Work and Position of This Study}
\label{sec:related}

\paragraph{Pocket heads on frozen trunks.}
zPocket \cite{zpocket} trains per-residue heads on the frozen pair
representation of co-folding models and reports a per-target AUPRC of
0.745, against 0.598 and 0.595 for the strongest retrained baselines
and 0.477 for P2Rank. Its heads form a capacity ladder, from a
single-residue readout through pooled pair features and axial mixing
to Pairformer-style blocks. We adopt its pooling rather than devising
another; the architectural comparison itself is not run here, for the
reason given in Appendix~\ref{sec:heads}.

\paragraph{Familiarity has already been shown to matter.}
zPocket also supplies PocketWatch, a temporally held-out benchmark
stratified by sequence similarity, structural similarity and
crypticity. Two findings bear on the axes defined here. Structural
similarity is the more informative leakage axis: per-target AUPRC is
roughly flat across sequence-similarity bins while varying strongly
with pocket coverage. And the variation is large, 0.803 for test
pockets with 80--100\% coverage against a structurally similar
pre-cutoff structure, against 0.327 for those below 20\%. That
coverage measure is, up to curation details, the quantity this study
calls \(S_p\). The familiarity effect this design was built to detect
is therefore not in doubt and we do not claim it.

\paragraph{What remains open.}
Those heads read a single representation, the trunk's final pair
tensor, so they answer where the pocket signal lives across
architectures and leave open where it lives across depth. A signal
present throughout the trunk but weakly decodable is a different
object from one that appears late, and the two imply different
interventions. Nor has the collapse on novel pockets been decomposed
into its family and geometry components. This study is accordingly a
depth-resolved diagnostic rather than a new pocket predictor: it
inherits the familiarity axes and the readout ladder, contributes the
onset machinery and its calibration, and reports where in a frozen
trunk pocket information becomes available under each regime. Where
our numbers restate what \cite{zpocket} has shown, we report them as
replication and say so.

\section{The Two-Axis Design}
\label{sec:design}

Each evaluation target carries two familiarity scores measured
against systems available before the trunk's training cutoff:
\(S_f\) for its protein family, from sequence clustering, and
\(S_p\) for its pocket geometry, from a pocket-level structural
measure rather than sequence identity, since the two axes must
measure different things for the design to separate them.
Thresholding both gives four regimes, \(FF\), \(FN\), \(NF\) and
\(NN\), where the first letter is family familiarity and the second
pocket familiarity.

The pre-registered primary contrast is \(FF\) against \(NN\). The
intermediate regimes are not decoration: \(FN\) and \(NF\) differ in
which axis is novel while closely sharing protein length, positive
rate and the system-level pocket measure, so a difference between
them is not attributable to those. The design therefore admits a
result in which the corner contrast is null while the intermediate
ones separate, which is what we find.

Both axes are also retained as continuous scores, so no conclusion
rests on the categorical form alone. The trunk stays frozen
throughout: the question is what is already accessible in the
representation, not what end-to-end adaptation could achieve.
Appendix~\ref{sec:design-full} gives the formal statement and
Appendix~\ref{sec:dataset-full} the construction.
The pre-registered hypotheses and kill criteria are in
Appendix~\ref{sec:hypotheses}.

\section{Dataset and the Two Axes}
\label{sec:dataset}

We use PLINDER \cite{plinder}, release 2026-09, which supplies both
familiarity axes as precomputed annotations. The frozen trunk is
Boltz-1 \cite{boltz1}, whose structural training cutoff of
\textbf{2021-09-30} does the load-bearing work: the reference pool
against which familiarity is measured is everything released on or
before it, and evaluation targets are systems released strictly
after. The same cutoff is used by AlphaFold~3, Chai-1 and Protenix,
so leakage-controlled splits built for those models carry over.

We first attempted Boltz-2 \cite{boltz2}, whose distributed
checkpoint lacks 128 parameters its own model class expects: a layer
norm inside the attention module of every one of its 64 trunk
blocks. Missing parameters are randomly initialised, so the model
runs, representations can be read from it, and every one of them is
partly noise. We record this because the failure is silent, and
because a depth-resolved analysis is exactly the kind of study it
would quietly invalidate.

Of 245{,}732 post-cutoff systems, 25{,}289 pass PLINDER's composite
validation flag and carry a pocket; those are the targets. The
reference pool is deliberately unfiltered, since the question
familiarity asks is what the trunk could have been exposed to.

\(S_f\), protein-family familiarity, comes from PLINDER's sequence
clustering joined through each system's own receptor chains, and is
established asymmetrically: one familiar chain suffices for
familiarity, while novelty requires every chain resolved and none
familiar. Treating a partly unresolved system as novel would
manufacture novel-family targets out of gaps in the cluster index.
98.2\% of chains carry an assignment and 98.1\% of systems have
every chain assigned.

\(S_p\), pocket familiarity, is the greatest \texttt{pocket\_qcov}
between a target pocket and any pre-cutoff pocket, aggregated over a
system's ligands by minimum and thresholded at 80. Two natural
alternatives proved unsuitable and a third metric starves the
novel-family cell; Appendix~\ref{sec:dataset-full} reports the
comparison, the threshold sweep that fixed 80 on cell balance before
any performance number existed, and the pool counts.

Two limitations travel with this. ``Novel'' means below a stated
similarity threshold against the pre-cutoff pool, not demonstrably
absent from pretraining: a post-cutoff release date establishes that
a structure was not in the training set, not that nothing resembling
it was. And the adopted metric has heavy ceiling mass, which limits
the resolution of the continuous analysis relative to the binary one.

\paragraph{Reproducibility.} The analysis used PLINDER release
2026-09 read through plinder-core 0.2.27. That package version
serves only release 2024-06 through its query interface and
substitutes it without warning, so the tables here are read as files
from the release directory. The per-ligand pocket-similarity scores
were computed in October 2026 and their generating script was not
retained; the rule they implement is recorded alongside the output. Code and results are at
\url{https://github.com/maryam-rahimi/Beyond_Pocket}.
\section{Representation-Depth and Pocket-Information Onset}

Pocket prediction from a final representation cannot say when the
information became accessible. Per-layer probing of co-folding
trunks is established practice --- \cite{feldman2026shortcut} probe
every Pairformer block of AlphaFold~3, and the same authors probe
its single and pair representations along the forward pass
\cite{feldman2026alphainterp} --- so our contribution is not the
extraction but the stratification: whether the depth at which pocket
information becomes decodable depends on familiarity.

We measure decodability independently at every block in each of the
four regimes, and report an onset rather than a peak. A maximum
chosen from forty-eight noisy estimates is inflated by the choosing,
and the question is when information becomes available rather than
where a curve happens to be highest. Reading only the final layer is
a questionable default in any case: intermediate layers carry more
usable signal across language and vision models
\cite{skean2025layer} and on molecular encoders
\cite{pinto2025intermediate}, and the advantage is largest under
distribution shift \cite{odonnat2026layer}.
Appendix~\ref{sec:onset-design} gives the full rationale, the
uncertainty treatment and the pre-declared contrasts.

%
%

\subsection{Null-Calibrated Onset}
\label{sec:onset}

For each evaluation regime \(q\) we construct a null distribution by
permuting pocket labels and refitting the entire probe pipeline.

\paragraph{Permutation scheme.}
Labels are permuted \emph{within} each protein, so every protein retains
its own positive rate and only the identity of the pocket-lining
residues is destroyed. Permuting across proteins would additionally
redistribute pocket size between them, making the null easier than the
real task for reasons unrelated to the probe and depressing the
threshold.

Replicate \(b\) uses the same within-protein permutation at every layer
and for every feature set. The null values for a replicate therefore
form a depth \emph{trajectory}
\[
N^{(b)}_{0},\;N^{(b)}_{1},\;\ldots,\;N^{(b)}_{L},
\]
rather than \(L+1\) unrelated draws. This is required rather than
convenient: the onset is defined by a run of consecutive exceedances, so
its error rate depends on the correlation between neighbouring layers,
and independent per-layer permutations would destroy exactly the
structure that must be measured.

\paragraph{Threshold.}
The threshold is computed separately at every layer, because the null
level reflects the probe's capacity to fit noise in that layer's
representation, which varies with depth. With \(B\) replicates and
level \(\alpha\),
\[
\tau_{\ell,q}
=
N^{(k)}_{\ell,q},
\qquad
k=\left\lceil (1-\alpha)(B+1)\right\rceil,
\]
the \(k\)-th smallest of the \(B\) null values at that layer. For
\(B=100\) and \(\alpha=0.05\) this is the 96th ordered value, with
exceedance probability at most \(5/101\approx0.0495\). An order
statistic is used rather than an interpolated quantile so that the
threshold is a value the null actually produced.

\paragraph{Onset.}
The pocket-information onset is
\[
\ell_q^*
=
\min
\left\{
\ell :
M_{\ell+j,q}>\tau_{\ell+j,q}
\;\;\text{for}\;\;
j=0,\ldots,4
\right\},
\]
so the signal must exceed the calibrated threshold \emph{at its own
layer} for five consecutive representation stages. The five stages are
adjacent layers \(\ell,\ldots,\ell+4\); a run of five non-adjacent
measurements spans more depth and is a weaker requirement, so the
analysis is run only on a complete, unit-spaced layer set.

The persistence value is not chosen arbitrarily. It is fixed by the
synthetic calibration in Section~\ref{sec:onset_validation}, which shows
that a three-stage requirement admits a severe early false positive
under transformer-like depth correlation.

\paragraph{Empirical false-positive rate.}
Because every replicate's per-layer score is retained, the rule's error
rate is measured on the study's own null rather than inferred from the
synthetic calibration. Each null trajectory \(b\) is tested against a
threshold built from the other \(B-1\) replicates,
\[
\tau^{(-b)}_{\ell,q},
\]
and we report the fraction of null trajectories that produce an onset.
The leave-one-out construction is necessary: a trajectory tested against
a threshold it helped raise is an in-sample comparison and understates
the rate. The observed curves are tested against the full-\(B\)
threshold; the leave-one-out thresholds are used only for the audit.

\subsection{Accumulation Depth as a Secondary Statistic}
\label{sec:l90}

An onset is undefined when a regime already exceeds its threshold at the
first measurable layer. We therefore pre-declare a second depth
statistic, fixed before any regime other than FF had been inspected:
\[
\ell_{90,q}
=
\min
\left\{
\ell :
M_{\ell,q}
\ \ge\
M_{0,q}+0.9\left(\max_{\ell'} M_{\ell',q}-M_{0,q}\right)
\right\},
\]
the first layer reaching 90\% of the gain available \emph{above the
first layer's own value}. Normalising by the gain rather than by
absolute height is what makes it informative in this setting: a
fraction-of-absolute statistic would be dominated by whatever the first
layer already provides.

\(\ell_{90}\) is secondary. The null-calibrated onset remains the
primary endpoint and both are reported regardless of outcome. The two
answer different questions, when the signal becomes detectable, and
when most of the depth-dependent gain has accumulated, and the second
remains defined when the first cannot resolve. Because \(\ell_{90}\) is
measured against the observed maximum, it inherits whatever noise placed
that maximum, and is an accumulation depth relative to the peak seen
rather than an estimate of a latent saturation point.

%
%

\subsection{Pre-Registration of the Persistence Rule by Synthetic Calibration}
\label{sec:onset_validation}

The onset rule defined above is a measuring instrument, and its error
properties cannot be established on the data it is meant to analyse: on real
representations the true onset is the unknown quantity being estimated, so a
recovered onset cannot be scored against anything. We therefore fix the
rule's free parameters in advance on synthetic representations where the
injected onset is known by construction, and report the rule's behaviour
there before applying it to any frozen trunk.

\paragraph{Construction.}
We generate \(n=200\) synthetic proteins of 40--90 residues across
\(L+1=65\) stages of width 32, with about 12\% of residues labelled
pocket in one contiguous run. A pocket signal is added along a fixed
random direction with per-stage strength \(g_\ell\), zero before
\(\ell^{\mathrm{true}}\) and rising smoothly after, so any earlier
reported onset is unambiguously a false positive. Two properties of
the real setting are reproduced: partitions are formed over proteins
rather than residues, and the null permutes labels within each
protein, preserving its label count.

Adjacent blocks of a transformer trunk are residual updates of one
another, so their representations are correlated, whereas
independently drawn synthetic stages are not. Since the persistence
requirement asks for consecutive stages above \(\tau\), and
correlated stages make a chance run more likely, this is the
property the rule must be exposed to. Stage noise is therefore an
AR(1) process in depth, \(\epsilon_\ell = \rho\epsilon_{\ell-1} +
\sqrt{1-\rho^2}\,\eta_\ell\), evaluated at \(\rho=0\) and
\(\rho=0.9\), with the signal added on top so
\(\ell^{\mathrm{true}}\) remains exactly true.

\paragraph{Result.}
Table~\ref{tab:onset_recovery} reports recovery under four injected
scenarios. A strong signal is recovered exactly in every configuration, and
the signal-free null is correctly reported as unresolved in every
configuration. The rule's failure is confined to the weak-signal regime, and
there it depends entirely on depth correlation: with independent stages the
weak onset is recovered two stages late, which is the benign direction, while
at $\rho=0.9$ the same signal yields a recovered onset of stage~1 against an
injected onset of stage 20, an error of $-19$ stages. Inspection of the underlying curve shows why:
its range, $0.1218$--$0.1636$, sits almost entirely against
$\tau=0.1317$, and correlated noise supplies three consecutive crossings
near the input embedding.

Raising the persistence requirement from three consecutive stages to five
removes this failure without cost to the other scenarios: at $\rho=0.9$ the
weak onset is recovered exactly, while the strong scenarios remain exact and
the null remains unresolved. We therefore pre-register
\[
\ell_q^{*}
=
\min\left\{\ell :
M_{\ell+j,q} > \tau_q \;\;\text{for}\;\; j=0,\ldots,4\right\},
\qquad \alpha = 0.05,
\]
as the onset rule used throughout the remainder of this study, fixed before
any frozen representation was extracted.


\begin{table}[h]
\centering
\caption{Onset-rule validation on synthetic representations with known
ground truth. Columns are rule configurations; rows are injected scenarios.
\emph{True} is the stage at which signal is first injected; cells give the
recovered onset with signed error. \texttt{unres.} denotes no onset
detected, the correct outcome for the null scenario. Bold marks an onset
reported before the signal exists.}
\label{tab:onset_recovery}
\begin{tabular}{l c c c c}
\toprule
Scenario & True & $\rho$=0.0, p=3 & $\rho$=0.9, p=3 & $\rho$=0.9, p=5 \\
\midrule
clean & 20 & 20 (+0) & 20 (+0) & 20 (+0) \\
early & 5 & 5 (+0) & 5 (+0) & 5 (+0) \\
weak & 20 & 22 (+2) & \textbf{1 ($-$19)} & 20 (+0) \\
null & -- & \texttt{unres.} & \texttt{unres.} & \texttt{unres.} \\
\bottomrule
\end{tabular}
\end{table}

\paragraph{Interval width carries information the point estimate does not.}
Table~\ref{tab:onset_ci} reports the protein-level bootstrap interval for
each recovered onset. The two quantities fail independently. Under
$\rho=0.9$ with three-stage persistence, the \emph{clean} scenario recovers
the injected stage exactly while its interval spans 21 of 65 stages, with a
lower bound at stage~0, the instrument was already degraded in a scenario
whose point estimate looked like a clean pass. Raising persistence to five
collapses that interval to a width of one. The weak scenario behaves
differently again: its interval remains 22 stages wide in every
configuration, including the one in which its point estimate is exactly
correct. We therefore report onset and interval together throughout, and
treat a recovered onset whose interval spans a large fraction of depth as
unresolved in substance even where the rule returns a value.


\begin{table}[h]
\centering
\caption{Protein-level bootstrap interval for the recovered onset, by
scenario and rule configuration, with width $w$ in stages (model depth is
65 stages). Width and point accuracy fail independently, so both are
reported. Bold marks an interval whose lower bound reaches stage~0, i.e.\
the bootstrap cannot exclude the signal being present from the input
embedding onward. The null scenario has no interval because no onset is
detected.}
\label{tab:onset_ci}
\begin{tabular}{l c c c}
\toprule
Scenario & $\rho$=0.0, p=3 & $\rho$=0.9, p=3 & $\rho$=0.9, p=5 \\
\midrule
clean & [14, 20] $w$=6 & \textbf{[0, 21] $w$=21} & [20, 21] $w$=1 \\
early & [5, 5] $w$=0 & \textbf{[0, 5] $w$=5} & [1, 5] $w$=4 \\
weak & \textbf{[0, 23] $w$=23} & \textbf{[0, 22] $w$=22} & \textbf{[0, 22] $w$=22} \\
null & -- & -- & -- \\
\bottomrule
\end{tabular}
\end{table}

\paragraph{A magnitude difference can masquerade as a timing difference.}
The depth contrast $\Delta_{\mathrm{onset}} = \ell_{NN}^{*} - \ell_{FF}^{*}$
is read as a statement about \emph{when} pocket information becomes
accessible. A weaker signal, however, crosses a fixed threshold later even
when it begins at the same stage, and the $NN$ regime is the regime expected
to be weaker, so the confound points in the same direction as the
hypothesis. To bound this, we run paired synthetic arms that share proteins,
partition, noise realisation and signal direction, and differ only in gain,
with onset held identical. Any nonzero contrast recovered under that
condition is delay purchased entirely with signal strength.

This quantity is a sensitivity benchmark, not a correction term. Its
magnitude depends on the gain ratio, ramp shape and signal-to-noise level
injected synthetically, none of which are calibrated against real
representations, so it is not an estimand and is not subtracted from any
measured $\Delta_{\mathrm{onset}}$. It is reported alongside the measured
contrast, and a timing interpretation is offered only where the measured
contrast exceeds the delay a pure magnitude difference produces under the
matched control.

\paragraph{Scope.}
This calibration fixes the rule's parameters and documents the regime in
which it fails; it does not establish that the rule behaves identically on
real co-folding representations, whose correlation structure, effective
dimensionality and signal strength are not known in advance to match the
synthetic construction. Its role is the one the pre-registration requires:
the persistence value and $\alpha$ used below were chosen on data with known
ground truth, before any real representation was seen, and not adjusted
afterwards.

%
%

\section{Results}
\label{sec:results}

\subsection{Pocket information is linearly decodable, and most of what
depth adds arrives early}

\begin{figure*}[t]
\centering
\includegraphics[width=\textwidth]{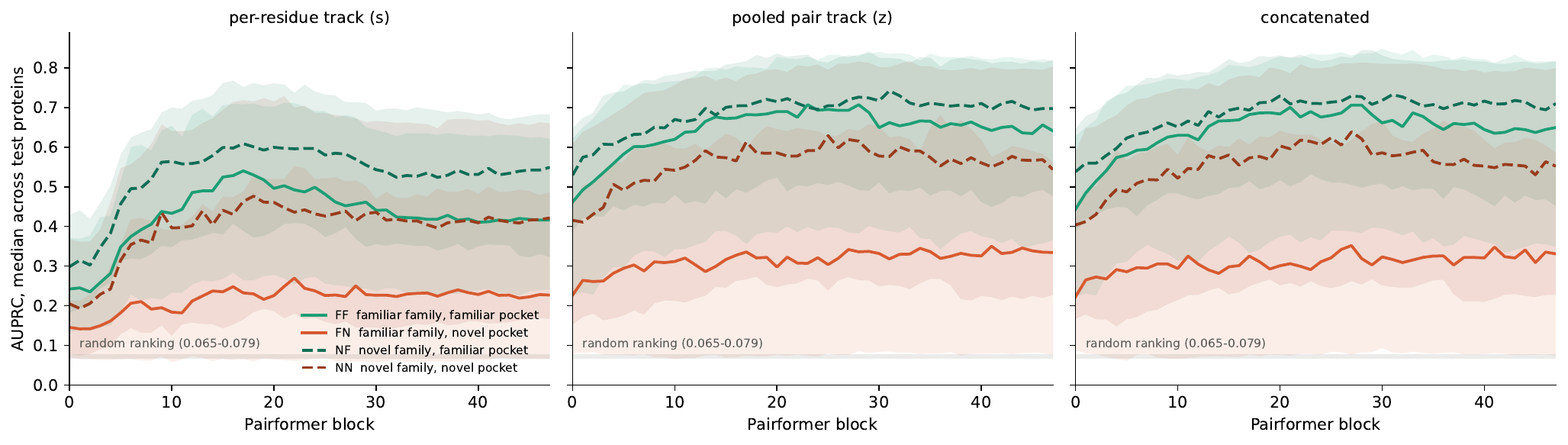}
\caption{\textbf{A structurally novel pocket on a familiar protein is
the hardest case.} \(FN\) sits below every other regime at every
depth and in every feature set, including below \(NN\), which is
novel on both axes. Colour encodes pocket familiarity, line style
protein-family familiarity. Curves are median AUPRC across 534 test
proteins; bands are the interquartile range across proteins, not a
confidence interval. The grey band spans the four regimes'
random-ranking baselines.}
\label{fig:curves}
\end{figure*}

Figure~\ref{fig:curves} shows the median held-out AUPRC against
Pairformer depth for each familiarity regime and each feature set, over
534 test proteins drawn from 1{,}820 extracted systems. Every curve
rises well above its random-ranking baseline: at its peak the weakest
regime reaches 3.4 times the rate a random ordering of residues would
achieve and the strongest 9.9 times
(Table~\ref{tab:curve-summary}). Pocket membership is therefore
linearly decodable from the frozen trunk, which this study assumed and
had not established.

The gain is concentrated in the first third of the trunk. Across the
twelve curves the layer at which 90\% of the gain available above layer
0 has accumulated lies between 14 and 27
(Table~\ref{tab:curve-summary}), and beyond roughly layer 20 the
measured AUPRC changes little. With 48 blocks available, most of the
depth-dependent improvement a linear probe can read is complete
before the halfway point.

The random-ranking baseline is not common to the four regimes: it is
each regime's own positive rate, and those range from 0.065 to 0.079.
Figure~\ref{fig:curves} draws the span rather than a single line, and
fold-over-baseline is reported alongside raw AUPRC for the same
reason.

\subsection{The hardest regime is a familiar family with a novel
pocket}

The cleanest comparison the design affords is \(FN\) against \(NF\).
The two differ in which axis is novel while agreeing on protein
length, positive rate, the system-level pocket measure and median
alignment depth, so a difference between them is not a difference in
evaluation geometry. Their median AUPRC differs by a factor of 2.6,
0.230 against 0.601 at the common evaluation layer, and the contrast
survives Holm correction over all eighteen reported contrasts in
every feature set (Table~\ref{tab:holm}). A structurally novel pocket
on a well-studied protein is harder to read than a familiar pocket on
an unfamiliar one.

It is also harder than a target novel on both axes. Ordering the four
regimes by peak AUPRC on the primary feature set gives \(NF\) 0.609,
\(FF\) 0.540, \(NN\) 0.476 and \(FN\) 0.270. Those are peaks, each
regime at the layer where its own curve is highest, which differ by a
few blocks; every contrast reported here is read instead at a single
common layer, so the two cells being compared come from the same
depth. The same ordering holds in all three feature sets
(Table~\ref{tab:curve-summary}).

The two axes therefore do not contribute equally, and not in the
direction a memorisation account predicts. Averaging over the family
axis, the familiar-pocket regimes exceed the novel-pocket regimes by
0.20 in AUPRC; averaging over the pocket axis, the
novel-\emph{family} regimes exceed the familiar-family ones by 0.14.
The first of these is established \cite{zpocket} and we report it as
a replication; the second runs opposite to what familiarity with a
protein family would suggest, and the regime-level contrast that
carries it does not survive correction (Table~\ref{tab:holm}).

This differs from what MIRAGE reports for binding affinity, where
protein-family support is the dominant measured axis \cite{mirage}.
The two results are not in conflict: affinity depends on the
chemistry of a particular family's site, whereas pocket membership is
closer to a shape property that may transfer across folds. Which
memorisation channel dominates appears to depend on what the model is
asked for, though the comparison is between different endpoints and
different layers, so it suggests rather than establishes this.

Four further checks are in Appendix~\ref{sec:results-full} and none
overturns this. The \(FN\)--\(NF\) gap is not an artefact of
evaluation geometry: the two cells agree on protein length, pocket
size and positive rate, so a regression on those covariates leaves
the regime difference intact. It is not an artefact of the pocket
metric: regrouping the same scores under a different similarity
measure preserves the ordering, though it shrinks the gap. It is not
explained by alignment depth, on which the two cells are
near-identical while differing 2.6-fold in score. And it is not
explained by what the cells contain: \(FN\) holds fewer cofactor and
more covalent ligands than \(NF\), but covalency is associated with
lower decodability across all four regimes and least so in \(FN\),
so the composition difference is a candidate confounder rather than
a mechanism.

\subsection{The hard regime is lowest wherever the boundary is drawn}

The four cells exist because a line was drawn at a particular value of
pocket coverage, so a finding that appears only near that value would
be a finding about the line. Holding the metric and the aggregation
fixed and moving only the threshold separates the two, which varying
the metric cannot.

\(FN\) has the lowest median AUPRC at every threshold tested and in
every feature set (Table~\ref{tab:threshold-sensitivity},
Figure~\ref{fig:threshold}). The size of its deficit varies
considerably --- a permissive boundary leaves a small and extreme
\(FN\), a strict one a large and milder --- but its position does not.

The rest of the ordering is not stable. \(NN\) rises steeply with the
threshold while \(FF\) falls beyond the primary value, and on the
per-residue track \(NF\) and \(FF\) converge at the strictest
boundary. We therefore report the position of \(FN\) as robust to the
operationalisation and the relative order of the other three as
specific to it.

\begin{table}[t]
\centering
\caption{Median AUPRC by regime as the pocket-familiarity threshold moves, with the metric held at \texttt{sp\_qcov} and the aggregation unchanged. Cell sizes vary with the threshold by construction: a strict boundary leaves few familiar-pocket systems and a permissive one few novel-pocket ones. Medians over fewer than 30 systems are shown in parentheses and are not used to establish an ordering. The same regime is lowest at every threshold in every feature set. These are regroupings of one probe's scores, not a re-run of the study at each threshold.}
\label{tab:threshold-sensitivity}
\begin{tabular}{llrrr}
\toprule
Features & Cell & \(\tau=50\) & \(\tau=80\) & \(\tau=100\) \\
\midrule
\texttt{s} & FF & 0.452 & 0.530 & 0.455 \\
 & FN & 0.079 & 0.230 & 0.365 \\
 & NF & 0.589 & 0.601 & 0.447 \\
 & NN & 0.219 & 0.476 & 0.562 \\
\addlinespace
\texttt{z} & FF & 0.607 & 0.695 & 0.637 \\
 & FN & 0.091 & 0.303 & 0.442 \\
 & NF & 0.709 & 0.704 & 0.727 \\
 & NN & 0.372 & 0.629 & 0.684 \\
\addlinespace
concat. & FF & 0.594 & 0.705 & 0.617 \\
 & FN & 0.085 & 0.353 & 0.433 \\
 & NF & 0.720 & 0.724 & 0.737 \\
 & NN & 0.341 & 0.637 & 0.692 \\
\bottomrule
\end{tabular}
\end{table}

\subsection{Which contrasts survive resampling}

Every comparison above is a point estimate on 122 to 152 test
proteins. We attach percentile intervals by resampling
receptor-family clusters with replacement, 2000 times, taking all of
a cluster's proteins when one is drawn, because proteins within a
family share sequence and fold and resampling them individually
would count near-duplicates as independent evidence. Each contrast
is computed inside the replicate. The intervals carry no correction
for the number of comparisons, so we also test them: each contrast
gets a two-sided p-value from the same resamples, re-centred on its
own estimate so the replicates describe a world in which the
contrast is zero, and Holm's correction is applied across all
\HolmFamily\ of them. The family was fixed at all \HolmFamily\
before the result; narrowing it to the contrasts involving one
regime after seeing which regime separates would be selection by
another route.

\HolmNFocusSurvive\ of the \HolmNFocus\ contrasts placing \(FN\)
below another regime survive at \(\alpha=\HolmAlpha\), with the
largest adjusted p-value among them \HolmWorstFocus. Of the
\HolmNOther\ contrasts not involving \(FN\), \HolmNOtherSurvive\
survive (Table~\ref{tab:holm}). On the primary feature set the three
are \(+0.300\), \(+0.371\) and \(+0.246\), so \(FN\) sits below each
of the other three regimes including the one novel on both axes.
\HolmNAtFloor\ raw p-values equal \HolmFloor, the smallest
\HolmNBoot\ replicates can produce, so the correction is
conservative for those.

Two contrasts contain zero before correction and survive none after.
One is \(NF\)--\(FF\), the apparent advantage of novel families over
familiar ones, which we do not interpret. The other is the
pre-registered primary contrast \(FF\)--\(NN\), \(+0.053\)
\([-0.096, +0.169]\) on the primary feature set and similar
elsewhere: the comparison the study was designed around returned
nothing. \(NF\)--\(NN\) excludes zero in two feature sets before
correction and survives in none, so we report it as unresolved.

\begin{table}[t]
\centering
\caption{The eighteen regime contrasts with Holm-Bonferroni correction. The multiplicity family is all 18 contrasts across all three feature sets, fixed before the result rather than narrowed to the contrasts that separate. Intervals are percentile intervals from 2000 resamples of receptor-family clusters; p-values are two-sided and null-centred on the same resamples. All 9 contrasts placing \(FN\) below another regime survive, in every feature set, and no other contrast does. 7 raw p-values equal \(1/(2000+1)\), the smallest value this many replicates can produce; their true values may be smaller, so the correction is conservative for them.}
\label{tab:holm}
\begin{tabular}{l l r l r r c}
\toprule
Features & Contrast & Estimate & 95\% interval & $p$ & $p_{\mathrm{Holm}}$ & Survives \\
\midrule
\texttt{s} & \(FF\)--\(FN\) & +0.300 & [+0.115, +0.403] & 0.0005 & 0.0090 & yes \\
\texttt{s} & \(NF\)--\(FN\) & +0.371 & [+0.211, +0.470] & 0.0005 & 0.0090 & yes \\
\texttt{s} & \(NN\)--\(FN\) & +0.246 & [+0.070, +0.374] & 0.0010 & 0.0110 & yes \\
\texttt{s} & \(NF\)--\(NN\) & +0.124 & [+0.018, +0.234] & 0.0255 & 0.2294 & no \\
\texttt{s} & \(FF\)--\(NN\) & +0.053 & [-0.096, +0.169] & 0.4323 & 1.0000 & no \\
\texttt{s} & \(NF\)--\(FF\) & +0.071 & [-0.015, +0.188] & 0.1714 & 1.0000 & no \\
\midrule
\texttt{zpool} & \(FF\)--\(FN\) & +0.392 & [+0.237, +0.504] & 0.0005 & 0.0090 & yes \\
\texttt{zpool} & \(NF\)--\(FN\) & +0.401 & [+0.281, +0.534] & 0.0005 & 0.0090 & yes \\
\texttt{zpool} & \(NN\)--\(FN\) & +0.326 & [+0.109, +0.479] & 0.0005 & 0.0090 & yes \\
\texttt{zpool} & \(NF\)--\(NN\) & +0.075 & [-0.017, +0.250] & 0.3928 & 1.0000 & no \\
\texttt{zpool} & \(FF\)--\(NN\) & +0.066 & [-0.071, +0.231] & 0.4938 & 1.0000 & no \\
\texttt{zpool} & \(NF\)--\(FF\) & +0.009 & [-0.039, +0.120] & 0.8321 & 1.0000 & no \\
\midrule
\texttt{both} & \(FF\)--\(FN\) & +0.352 & [+0.215, +0.519] & 0.0005 & 0.0090 & yes \\
\texttt{both} & \(NF\)--\(FN\) & +0.371 & [+0.298, +0.552] & 0.0005 & 0.0090 & yes \\
\texttt{both} & \(NN\)--\(FN\) & +0.285 & [+0.097, +0.483] & 0.0015 & 0.0150 & yes \\
\texttt{both} & \(NF\)--\(NN\) & +0.086 & [+0.004, +0.258] & 0.2604 & 1.0000 & no \\
\texttt{both} & \(FF\)--\(NN\) & +0.068 & [-0.084, +0.234] & 0.4248 & 1.0000 & no \\
\texttt{both} & \(NF\)--\(FF\) & +0.019 & [-0.039, +0.146] & 0.6037 & 1.0000 & no \\
\bottomrule
\end{tabular}
\end{table}

\subsection{Onset, and why it does not discriminate}
\label{sec:results-onset}

The null calibration refits the entire pipeline on labels permuted
within each protein, one hundred times at each of the 48 layers, with
replicate $b$ holding the same permutation across depth so that the
resulting values form null trajectories rather than independent draws.
The threshold $\tau_{\ell,q}$ is the finite-sample order statistic of
those values at each layer.

Two things follow from it.

\paragraph{The thresholds are nearly flat with depth.} Across all 48
layers $\tau$ varies by about $0.003$ within each regime: $0.094$ to
$0.097$ in \(FF\), $0.083$ to $0.086$ in \(FN\) and \(NN\), $0.088$ to
$0.092$ in \(NF\). The pipeline's capacity to fit shuffled labels
therefore barely changes between the first Pairformer block and the
last, which is not obvious in advance and means a single threshold
would have been a fair approximation here. Each $\tau$ sits roughly
$0.016$ above its regime's positive rate, which is the margin the
positive rate alone would have missed.

\paragraph{Every curve clears its threshold at every layer.} The
pre-registered onset --- the first of five adjacent layers exceeding
$\tau$ at their own depth --- is layer 0 in all four regimes, with 48
of 48 layers above threshold
(Table~\ref{tab:onset-observed}). $\Delta_{\mathrm{onset}} =
\ell^*_{NN} - \ell^*_{FF} = 0$.

This is a property of the data rather than a failure of the
estimator. Pocket membership is already decodable above chance where
the trunk begins, so the layer at which it \emph{first} becomes
decodable has no room to vary and cannot separate the regimes. We
report the primary endpoint as uninformative rather than reinterpreting
it.

\paragraph{The accumulation depth carries the comparison instead.}
$\ell_{90}$, the first layer reaching 90\% of the gain available above
layer 0, was pre-declared as a secondary statistic for this case, while
only the \(FF\) regime had been inspected. On the primary feature set
it gives $NF$ 14, $FF$ 15, $NN$ 17 and $FN$ 22.

One part of that ordering is stable and the rest is not. \(FN\) is the
latest regime in all three feature sets 22, 27 and 26 for \(s\),
pooled \(z\) and the concatenation so a novel pocket in a familiar
family takes consistently longer to accumulate its gain. The relative
order of the other three changes between feature sets: \(NF\) is
earliest for \(s\), \(NN\) for pooled \(z\), \(FF\) for the
concatenation. We report the \(FN\) result and do not interpret the
remaining ordering.

\paragraph{The persistence rule controls error on this study's own
null.} Section~\ref{sec:onset_validation} fixed the persistence at five
using synthetic curves with a known onset, where a three-stage
requirement erred by nineteen stages. That is a simulation result, and
the same question can be asked of the real representations: how often
does each rule fire on a null trajectory built from them? Testing each
trajectory against a threshold reconstructed from the other $B-1$
replicates gives a rate of $0.04$ to $0.07$ at five stages, against a
nominal $\alpha = 0.05$, and $0.12$ to $0.17$ at three
(Table~\ref{tab:onset-fpr}).

The calibrated rule therefore controls error near its nominal level on
the data it is applied to, and the naive rule exceeds it by between two
and three times. The synthetic failure is reproduced in kind on real
depth trajectories, at a magnitude the simulation could not have
predicted. A single-layer criterion fires on roughly seven null
trajectories in ten.

\section{What Is Not Established}

Four things. The \(\ell_{90}\) ordering carries no intervals: the
bootstrap applies to the cell medians rather than to the depth at
which a curve reaches a fraction of its own range. The family-axis
difference, \(NF\) above \(FF\) by 0.069 on the primary feature set,
is small and does not survive correction, and we do not interpret
it. Whether the representation can express pocket membership at all,
separately from whether a probe recovers it, would need a
reconstruction floor; at this width the finite sample is effectively
full-rank, so the floor separates nothing. And whether a richer
readout recovers more in the hard regime is the capacity question of
Appendix~\ref{sec:heads}, where the one nonlinear head we fitted
overfitted within an epoch.

All of this is one trunk. Whether the ordering is a property of
Boltz-1 or of co-folding representations generally is not addressed
by these data.

\section{Where the Difference Originates}

The depth curves show the regime difference already present at the
output of the first Pairformer block, and the remaining 47 preserve
rather than create it. We measure after that block, not before it, so
this bounds where the difference can have arisen rather than locating
it: whatever produced it had done so by the end of block one. That
rules out one class of explanation and leaves another.

It rules out blocks two through 48. Whatever they compute, they are
not what makes one regime harder than another, so a remedy that reads
a different layer cannot help.

It does not identify what upstream means. The difference enters
through the input representation, the alignment, or what the model
learned during pretraining, and these data do not separate them.
Alignment depth is excluded; alignment diversity is not tested.

And it leaves open whether a readout with more capacity could recover
what a linear probe cannot, which is a question about accessibility
rather than about origin. Section~\ref{sec:heads} says why that was
not answered here.

\section{Statistical Analysis}

All inference is at the protein level, never the residue level:
residues within a protein are not independent replicates, and
treating them as such would give intervals far too narrow.

Uncertainty comes from resampling receptor-family clusters rather
than proteins. Proteins within a family share sequence and fold, so
resampling them individually counts near-duplicates as independent
evidence; cluster resampling roughly doubles the interval widths.
Contrasts are computed inside each replicate rather than from
separately resampled estimates.

Thresholds for the quadrant analysis were fixed before any
performance number existed, and the primary contrasts were declared
in advance. The eighteen regime contrasts carry Holm correction over the whole
family, fixed before the result. The composition checks and the
covalent analyses do not, and the paper says so at each point, so
there an interval excluding zero is an interval rather than a test at
a stated level.

\section{Discussion}

The methodological result is the one that travels. The rule for
reading a depth curve is wrong in a way per-layer significance
testing cannot reveal, because the problem is between the layers.
The fix is borrowed from neuroimaging rather than invented, but
nobody had checked whether depth curves needed it, and they do.

What the estimator then found was not what the design was built to
find. Pocket information is decodable at the first block in every
regime, so the onset, the pre-registered endpoint, has nothing to
distinguish. That is a property of the data rather than a failed
measurement, and it is worth stating plainly because a study that
reinterprets its endpoint after seeing the result has stopped
measuring and started arguing.

The empirical finding is the hard cell. A familiar protein family
carrying a structurally novel pocket is harder than a target novel on
both axes, and that ordering survives every threshold tested, controls
for protein length and median alignment depth, and an alternative
definition of pocket familiarity. It is not explained by the composition
difference that first suggested an explanation: covalency is
associated with lower decodability across the board, but least so in
the regime whose composition raised the question.

Fitted continuously, the family axis is not null. More pre-cutoff
relatives go with lower decodability, which is the opposite of what a
memorisation account predicts and invisible when the axis is split in
two. We are cautious about it: the range sampled is narrow, and support
may track difficulty rather than cause it.
Three limits bound all of this. The result is one trunk. The deficit
is bounded to the input or the first block and not resolved within
that: input representation, alignment and pretraining are not
separated here.
And
what we measure is linear recoverability under a fixed readout, which
is narrower than whether the information is present at all.

The experiments those limits name are the obvious next ones. A second
co-folding trunk would say whether any of this is about Boltz-1 or
about these representations generally. Alignment diversity rather than
depth is the quantity the literature points at, and is computable from
the alignments already cached. And the alignment's contribution can be
tested directly: re-extracting with single-sequence inputs and
comparing the regimes would show which of them the MSA was helping,
which is a sharper question than whether its depth correlates with
anything. An extraction retaining the pair tensor would make the
capacity question answerable at the same time.

\section{Conclusion}

Requiring three consecutive layers above a null-calibrated threshold
reports an onset nineteen stages early on synthetic curves with known
ground truth, and fires on 12 to 17 per cent of null trajectories
built from real representations against a nominal 5 per cent; the
calibrated five-stage rule fires on 4 to 7. The failure follows from
correlation between neighbouring layers and applies wherever
depth-resolved probing is done.
Applied to pocket information in a frozen co-folding trunk, the
estimator returns an onset of zero in all four familiarity regimes:
pocket membership is already decodable above threshold at the first
Pairformer block, so the onset cannot distinguish them. We report that
rather than reinterpreting the endpoint. The pre-declared accumulation depth is latest for familiar families with structurally novel pockets in every feature set.

That regime is also the weakest. Its median AUPRC sits below each of
the other three with bootstrap intervals excluding zero, including
below the regime novel on both axes, and the gap survives controls for protein length, the system-level
pocket measure and alignment depth. The pre-registered
corner contrast returned nothing: its interval contains zero in all
three feature sets. The separation the design was built to detect is
there, but between the intermediate cells rather than the corners.


\clearpage

\bibliographystyle{plain}
\bibliography{references}

\clearpage
\appendix
\floatplacement{table}{H}
\floatplacement{figure}{H}

\section{Key Hypotheses}
\label{sec:hypotheses}

\subsection{Two-Axis Generalization Hypothesis}
The pre-registered primary hypothesis is
\[
P_{NN} < P_{FF},
\]
with the intermediate regimes determining whether any drop is driven by
family novelty, pocket novelty, or their interaction. We retain it as
stated. Section~\ref{sec:results} reports that its bootstrap interval
contains zero in all three feature sets, and that the separation
instead lies between \(FN\) and each of the other three regimes.

\subsection{Representation-Onset Hypothesis}

We test whether
\[
\ell_{NN}^*
>
\ell_{FF}^*.
\]
Section~\ref{sec:results-onset} reports both onsets as zero: pocket
membership exceeds the calibrated threshold at every layer in every
regime, so the contrast is not estimable from these data.
\section{Readout-Capacity Hierarchy}
\label{sec:heads}

This analysis is not reported here. The depth curves place the regime
difference at the first Pairformer block
(Section~\ref{sec:results}), which a readout operating on the trunk's
output cannot address, and the head comparison is left to future
work.

The design is zPocket's ladder \cite{zpocket} --- single-residue,
pooled pair, axial mixing, Pairformer-style --- adopted rather than
devised so that the comparison would have been a replication at depth
rather than a competing ladder. It is retained here because the
question remains open for a model whose regime difference arises
within the trunk, which this one's does not.

%
%
%

\subsection{A Gated Head, and the Gate as a Measurement}
\label{sec:gated-head}

The hierarchy above fixes each head's processing capacity in advance: a
per-residue readout never consults pair features, and a Pairformer-style
readout always does. Neither matches the expectation that pair information
matters unevenly  that a residue in a well-precedented site may be
identifiable from its own features alone, while one in an unfamiliar site
requires the context its neighbours supply.

We therefore add a fifth readout in which the balance is learned rather than
declared. Let \(u_i\) denote the single-feature readout for residue \(i\) and
\(v_i\) the pair-derived readout. The gated head computes
\[
g_i \;=\; \sigma\!\left(W\left[u_i \,;\, v_i\right] + b\right),
\qquad
\widehat{y}_i \;=\; g_i \odot v_i \;+\; (1-g_i)\odot u_i ,
\]
so that \(g_i \in (0,1)\) sets, per residue, how much of the prediction is
drawn from pair processing. The construction follows the gated fusion used to
combine a frozen protein language model with a specialised side module in
PTM-Mamba \cite{ptmmamba}, where a sigmoid-activated linear gate mixes two
embeddings rather than committing to either.

Two things recommend this head beyond the possibility that it performs well.

First, it subsumes the rest of the hierarchy. A gate saturated at zero
recovers the per-residue readout and one saturated at one recovers the pair
readout, so the comparison in Section~\ref{sec:gated-head} is not between
unrelated architectures but between a fixed choice and a learned one.

Second, and more useful diagnostically, \(g_i\) is itself a measurement.
Aggregated over residues and compared across familiarity regimes,
\[
\bar{g}_{q} \;=\; \frac{1}{n_q}\sum_{i \in q} g_i ,
\]
it reports how much pair processing the readout elects to use in regime
\(q\). The prediction that follows from the readout-interaction hypothesis is
\[
\bar{g}_{NN} > \bar{g}_{FF},
\]
that is, novel targets draw more on residue-pair context than familiar ones
do. This is a sharper statement than the performance contrast
\(\Delta_h^{NN} > \Delta_h^{FF}\), because it does not require the richer
head to \emph{win} in order to be informative: a gate that opens under
novelty while performance stays flat indicates that pair information is
being recruited and is not sufficient, which distinguishes a readout
limitation from a representation limitation more directly than accuracy
alone.

The gate is reported with the same protein-level resampling and seed
replication as the other heads, and its contrast across regimes is treated as
a secondary endpoint rather than a primary one, since it describes the
readout's behaviour rather than the task performance the study is organised
around.

%
%

\section{Dataset Construction in Full}
\label{sec:dataset-full}

The two familiarity axes defined above are abstract until instantiated on a
dataset with a fixed temporal cutoff. This section reports that
instantiation, the resulting evaluation pool, and the construction of both
familiarity axes. Two of the study's own kill criteria, whether all four regimes can be
populated and whether \(NN\) is large enough for stable evaluation,
are properties of the data alone and are therefore settled before any
representation is extracted.

\subsection{Dataset and Temporal Cutoff}

We use PLINDER \cite{plinder}, release 2026-09, which provides
protein--ligand interaction systems annotated with similarity measures at
the protein, pocket, interaction and ligand levels. It is the only resource
we are aware of that supplies both of our axes as precomputed annotations
rather than requiring them to be derived.

The frozen trunk is Boltz-1 \cite{boltz1}, chosen because its weights are
openly licensed, its Pairformer stack of 48 blocks exposes intermediate
representations, and, a criterion worth stating because it is not
automatic, its published weights load against its published code.

That last point decided the choice. We first attempted the newer Boltz-2
\cite{boltz2}, whose distributed checkpoint lacks 128 parameters that the
distributed
model class expects: a layer norm inside the attention module of every one
of its 64 trunk blocks. Missing parameters are randomly initialised, so the
model runs, intermediate representations can be read from it, and every one
of them is partly noise. No public commit of the repository matches the
checkpoint. We record this because the failure is silent, nothing in the output
announces it, and because a depth-resolved analysis is exactly the
kind of study it would quietly invalidate.

Boltz-1's structural training cutoff is \textbf{2021-09-30}. This date does
the load-bearing work in the design: the reference pool against which
familiarity is measured is everything released on or before it, and
evaluation targets are systems released strictly after it. The boundary is
treated as exclusive, because the model is reported as trained on structures
``up to'' that date, placing the boundary day itself inside training.

This is the same cutoff used by AlphaFold3, Chai-1 and Protenix, so
leakage-controlled splits built for those models carry over directly. It
also leaves more evaluation data than a later cutoff would: the newer
Boltz-2 cutoff of 2023-06-01 would have reduced the post-cutoff pool by
roughly a third.

\subsection{The Evaluation Pool}

Table~\ref{tab:pool} reports the pool. The temporal cutoff is not the
binding constraint: 245,732 systems postdate it, with no unparseable
release dates in the release.

\begin{table}[h]
\centering
\caption{Evaluation pool by trunk training cutoff. This study uses 2021--09--30; the 2023--06--01 column is shown for comparison and bounds what the alternative cutoff would cost in available data. The boundary is exclusive, so a system released on the cutoff date counts as pre-cutoff.}
\label{tab:pool}
\begin{tabular}{l r r}
\toprule
 & cutoff 2021-09-30 & cutoff 2023-06-01 \\
\midrule
PLINDER systems, total & \multicolumn{2}{c}{611,929} \\
pre-cutoff (reference pool) & 366,197 & 451,799 \\
post-cutoff (candidates) & 245,732 & 160,130 \\
unparseable release dates & \multicolumn{2}{c}{0} \\
\bottomrule
\end{tabular}
\end{table}

The binding constraint is quality rather than time. PLINDER's composite
validation flag holds for 18\% of systems, and requiring it of the targets
--- where pocket ground truth must be trustworthy --- reduces the pool
sharply:
\[
245{,}732
\;\xrightarrow{\text{validated}}\;
25{,}933
\;\xrightarrow{\text{has pocket}}\;
25{,}289 .
\]
Whether this filter is necessary, or whether a looser criterion would supply
adequate ground truth with substantially more data, is left open and
pre-registered as a decision to be made on the filter's definition rather
than on downstream performance.

The reference pool is deliberately \emph{not} filtered for quality. The
question familiarity asks is what the trunk could have been exposed to, and
Boltz-1 trained on the PDB rather than on a curated subset; restricting the
reference pool would understate exposure and inflate apparent novelty.

\subsection{Instantiating \texorpdfstring{\(S_f\)}{S\_f}}

Protein-family familiarity is taken from PLINDER's MMseqs sequence
clustering, joined through each system's own receptor chains. Two features
of that join are worth stating, because both admit a plausible shortcut that
biases the result toward the familiar regimes.

First, the join is chain-level rather than entry-level. A single PDB entry
can contain a chain belonging to a well-represented family alongside one
that does not, and a system whose receptor is only the latter would inherit
the former's familiarity if familiarity were resolved per entry. The effect
is directional: it inflates \(FF\) and depletes \(NN\), which is precisely
the regime the study exists to measure.

Second, familiarity and novelty are established asymmetrically:
\[
S_f(s)=
\begin{cases}
F, & \text{if any receptor chain of } s \text{ is known familiar},\\
N, & \text{if every receptor chain is resolved and none is familiar},\\
\text{unresolved}, & \text{otherwise}.
\end{cases}
\]
One familiar chain suffices to show the model has seen related structure,
but novelty is a claim about all of them and therefore requires complete
coverage. Treating a partially unresolved system as novel would manufacture
\(NN\) targets out of gaps in the cluster index. In this release the
question is small but not empty: 98.2\% of receptor chains carry a cluster
assignment and 98.1\% of systems have every receptor chain assigned.

\subsection{Constructing the Pocket Axis}
\label{sec:sp}

The pocket axis required more care than the family axis, and two natural
candidates proved unsuitable before a workable one was found. We report the
failures because both are available, plausibly named, and silent: each
yields a well-formed table rather than an error.

\paragraph{A representative-assignment table is not a similarity measure.}
PLINDER's \texttt{ligand\_\allowbreak pocket\_\allowbreak membership} assigns each system a
representative and is the obvious first candidate. Used as \(S_p\) it
labelled every target pocket-novel, emptying \(FF\) and \(NF\) entirely. The
cause is structural: the table is a near-identity deduplication rather than
a similarity clustering, with roughly nine in ten of its groups containing a
single system and most systems representing themselves. Under such a
grouping, sharing a representative with a pre-cutoff system is almost never
true, and the axis degenerates.

\paragraph{Domain annotations measure the wrong construct.}
The \texttt{system\_pocket\_CATH}, \texttt{SCOP2} and \texttt{Pfam} columns
group systems by the domain family a pocket sits in. SCOP2 is too sparse to
use; CATH and Pfam have workable group sizes but classify the protein family
rather than the binding site, which is close to what \(S_f\) already
measures from sequence. Using one as \(S_p\) produced a viable-looking
design whose \(FN\) cell held fewer than thirty receptor families --- the
two axes collapsing toward a single measurement, which is precisely the
condition the first kill criterion is written to detect. That scarcity was
an artefact of the proxy rather than a fact about the data, as the next
subsection shows.

\paragraph{The pairwise export supplies the right quantity.}
PLINDER's directed ligand-pair export provides pocket-level similarity
computed from per-residue alignments, reported as integer percentages. For
each candidate metric we define
\[
S_p(\ell)
=
\max_{j \,:\, d_j \le \text{cutoff}} \ \mathrm{sim}(\ell, j),
\]
the greatest similarity between target pocket \(\ell\) and any pocket
available before the trunk's training cutoff, excluding the target's own
system. This is continuous, pocket-level rather than chain-level, restricted
to the pretraining reference pool, and constructed from a different
measurement than \(S_f\).

\subsection{Choosing the Metric and the Aggregation}

Three metrics in the export are candidates, and
Table~\ref{tab:sp_metrics} reports their distributions.
Scores attach to ligands while \(S_p\) is a property of a system, so each
metric is shown per ligand and aggregated to the system by minimum and by
maximum; about one target in seven carries more than one ligand, so the
aggregation is a real choice rather than a formality.

\begin{table}[h]
\centering
\caption{Candidate pocket-similarity metrics, each taken as the maximum against the pre-cutoff pool. All are directed, pocket-level, and reported by PLINDER as integer percentages. Scores are stored per ligand, while $S_p$ is a system-level quantity, so each metric is shown three ways: the raw per-ligand distribution and the system-level distribution under minimum and maximum aggregation. The percentiles matter as much as the means, since a metric whose median sits at the ceiling can separate the binary quadrants while leaving the continuous familiarity regression with little resolution.}
\label{tab:sp_metrics}
\begin{tabular}{l l r r r r r r}
\toprule
metric & level & mean & 10\% & 25\% & median & 75\% & 90\% \\
\midrule
\texttt{pocket\_qcov} & per-ligand & 91.4 & 75 & 91 & 100 & 100 & 100 \\
 & system (min) & 90.7 & 72 & 90 & 97 & 100 & 100 \\
 & system (max) & 91.8 & 76 & 92 & 100 & 100 & 100 \\
\addlinespace
\texttt{pocket\_fident\_qcov} & per-ligand & 80.9 & 37 & 71 & 94 & 100 & 100 \\
 & system (min) & 79.8 & 34 & 69 & 92 & 100 & 100 \\
 & system (max) & 81.5 & 38 & 72 & 94 & 100 & 100 \\
\addlinespace
\texttt{pli\_qcov} & per-ligand & 69.9 & 0 & 53 & 82 & 100 & 100 \\
 & system (min) & 67.2 & 0 & 50 & 80 & 100 & 100 \\
 & system (max) & 72.1 & 0 & 59 & 85 & 100 & 100 \\
\addlinespace
\bottomrule
\end{tabular}
\end{table}

The distributions invite a conclusion the quadrant counts then overturn.
\texttt{pocket\_qcov} has the heaviest ceiling mass of the three, which
limits the resolution available to the continuous familiarity regression,
and on that basis \texttt{pocket\_fident\_qcov} looks preferable. Table
\ref{tab:sp_metric_choice} shows why it is not.

\begin{table}[h]
\centering
\caption{Family-cluster counts per quadrant by pocket metric, system-level aggregation and threshold. The weakest cell is the figure of merit, because the four-cell comparison is only as strong as its thinnest cell. The comparison is descriptive: it shows how metric choice and ligand aggregation affect four-cell support, and which definition to adopt follows from these numbers rather than being assumed.}
\label{tab:sp_metric_choice}
\begin{tabular}{l l r r r r r r}
\toprule
metric & agg. & thr. & FF & FN & NF & NN & weakest \\
\midrule
\texttt{pocket\_qcov} & max & 70 & 2,030 & 294 & 837 & 316 & \textbf{294} \\
\texttt{pocket\_qcov} & max & 90 & 1,809 & 808 & 532 & 666 & \textbf{532} \\
\texttt{pocket\_qcov} & min & 70 & 2,015 & 346 & 807 & 367 & \textbf{346} \\
\texttt{pocket\_qcov} & min & 90 & 1,778 & 924 & 495 & 716 & \textbf{495} \\
\texttt{pocket\_fident\_qcov} & max & 70 & 1,831 & 639 & 214 & 898 & \textbf{214} \\
\texttt{pocket\_fident\_qcov} & max & 90 & 1,405 & 1,332 & 70 & 998 & \textbf{70} \\
\texttt{pocket\_fident\_qcov} & min & 70 & 1,798 & 709 & 203 & 907 & \textbf{203} \\
\texttt{pocket\_fident\_qcov} & min & 90 & 1,349 & 1,423 & 69 & 999 & \textbf{69} \\
\texttt{pli\_qcov} & max & 70 & 1,655 & 1,203 & 500 & 745 & \textbf{500} \\
\texttt{pli\_qcov} & max & 90 & 1,290 & 1,743 & 268 & 943 & \textbf{268} \\
\texttt{pli\_qcov} & min & 70 & 1,598 & 1,315 & 455 & 781 & \textbf{455} \\
\texttt{pli\_qcov} & min & 90 & 1,201 & 1,817 & 232 & 953 & \textbf{232} \\
\bottomrule
\end{tabular}
\end{table}

Judged on the weakest cell, which is the right figure of merit since a
four-cell comparison is only as strong as its thinnest,
\texttt{pocket\_qcov} has the largest weakest cell at threshold 90 in
both aggregations, and \texttt{pli\_qcov} overtakes it at 70, while
\texttt{pocket\_fident\_qcov} is far behind throughout. The reason is specific
and worth stating: both alternatives are stricter about what counts as a
familiar pocket, \texttt{pocket\_fident\_qcov} by requiring the pocket
residues themselves to match and \texttt{pli\_qcov} by requiring the
interaction pattern to match. A protein from an unfamiliar family seldom satisfies either, so
\(NF\), novel family, familiar pocket, is the cell they starve.
At the permissive threshold that is not yet the binding constraint
for \texttt{pli\_qcov}, whose other cells are better populated, but
by 90 it is. That is the axis correlation the design must avoid,
reappearing by a different route than the domain annotations took.

Between the two aggregations the difference is small, which is what makes
the choice easy: we take the minimum over a system's ligands, the more
conservative reading of novelty, since a system containing one genuinely
novel pocket should not be called pocket-familiar on the strength of a
second, familiar one. We therefore adopt
\[
S_p(s) \;=\; \min_{\ell \in L_s} \ \max_{j \,:\, d_j \le \text{cutoff}}
\ \texttt{pocket\_qcov}(\ell, j).
\]

\subsection{Quadrant Occupancy}

Table~\ref{tab:sp_quadrants} reports occupancy across the threshold range,
in systems and in distinct receptor-family clusters. The cluster counts are
the ones that govern statistical weight, and they are the smaller numbers by
close to an order of magnitude.

\begin{table}[h]
\centering
\caption{Quadrant occupancy across candidate $S_p$ thresholds ($n=25,289$ targets). $S_p$ is the maximum \texttt{pocket\_qcov} against any pre-cutoff system, aggregated over a system's ligands by minimum. Each cell gives systems and, after the slash, distinct receptor-family clusters; the cluster count governs statistical weight. The thresholds 50, 70 and 95 are those PLINDER uses to build its own similarity graphs.}
\label{tab:sp_quadrants}
\begin{tabular}{r r r r r}
\toprule
threshold & FF & FN & NF & NN \\
\midrule
50 & 20,225 / 2,095 & 519 / 171 & 3,858 / 948 & 687 / 180 \\
70 & 19,648 / 2,015 & 1,096 / 346 & 3,113 / 807 & 1,432 / 367 \\
80 & 19,013 / 1,959 & 1,731 / 515 & 2,607 / 691 & 1,938 / 503 \\
90 & 17,003 / 1,778 & 3,741 / 924 & 1,729 / 495 & 2,816 / 716 \\
95 & 14,659 / 1,597 & 6,085 / 1,276 & 1,140 / 361 & 3,405 / 831 \\
100 & 10,950 / 1,338 & 9,794 / 1,629 & 594 / 222 & 3,951 / 942 \\
\bottomrule
\end{tabular}
\end{table}

All four regimes are populated at every threshold tested, and the weakest
cell exceeds a thirty-cluster planning target throughout --- the design does
not depend on a fortunate threshold choice. The two cells trade off
monotonically as the threshold rises: \(FN\) grows and \(NF\) shrinks, since
a stricter definition of pocket familiarity moves targets from the familiar
column to the novel one. \(NF\) is accordingly the scarcer cell at strict
thresholds and \(FN\) at permissive ones, with the best balance at 80,
where the weakest cell holds 503 receptor families.

The thresholds 50, 70, 95 and 100 are the values PLINDER uses to build its
own similarity graphs, so they carry justification external to this study;
80 and 90 are included to show the behaviour between them. We report the
full sweep rather than a single value, and any threshold adopted for the
primary analysis is fixed before performance numbers exist.

\subsection{What This Settles}

The instantiation above settles both data-dependent kill criteria. All four
regimes are populated at usable size, so the axes are not so correlated as
to prevent the design; and \(NN\) is large enough for pocket-level
evaluation. Neither conclusion required a model.

Two limitations travel with the result. ``Novel'' means below a stated
similarity threshold against the pre-cutoff pool, not demonstrably absent
from pretraining: a post-cutoff release date establishes that a structure
was not in the trunk's training set, not that nothing resembling it was.
The adopted metric's ceiling mass, while reduced by minimum
aggregation, still limits the resolution of the continuous analysis relative
to the binary one. The continuous fit is reported under
\texttt{pocket\_qcov} alone; the alternative metric starves \(NF\),
as the metric-choice table shows, so a sensitivity check under it
would compare a well-populated fit against a thin one.

Finally, what the depth analysis reads from the trunk is the output of each
Pairformer block, not a post-trunk transform of it. The onset reported is
therefore the stage at which pocket information becomes accessible
\emph{within the stack}, which is the intended object but is a narrower
claim than accessibility in whatever representation the structure module
ultimately consumes.

\paragraph{Reproducibility.} The analysis used PLINDER release
2026-09, read through plinder-core 0.2.27. That package version
serves only release 2024-06 through its query interface and
substitutes it without warning, so the tables here are read as files
from the release directory rather than fetched; a script that calls
the interface gets a different snapshot from the one this study used.
The per-ligand pocket-similarity scores were computed in October 2026
and their generating script was not retained. The rule they implement
is recorded alongside the output, and the scores themselves are
available, but regenerating them means re-implementing from that
record rather than re-running.

\section{Onset Design in Full}
\label{sec:onset-design}

\subsection{Why Onset Rather Than Peak}

Selecting
\[
\arg\max_\ell M_{\ell,q}
\]
is vulnerable to selection-induced inflation because the maximum is chosen from many noisy layer-wise estimates.

The scientific question is also different. We want to know when pocket information first becomes reliably available, not where a noisy curve happens to attain its largest observed value.

We therefore define an \emph{onset} rather than a peak.
Reading only the final layer is a questionable default in any case.
Intermediate layers carry more usable signal than final ones across
language and vision models \cite{skean2025layer} and in most
model--task pairings on molecular encoders
\cite{pinto2025intermediate}, and the advantage is largest under
distribution shift \cite{odonnat2026layer}, which is what the
familiarity regimes here construct.

The construction we adopt is not new as statistics. Cluster-based
permutation inference in neuroimaging builds null distributions by
permuting labels while preserving the time structure, and requiring
consecutive significant points to control the family-wise error rate
is standard there \cite{jeanrichard2020transients}.
\cite{rousselet2025onsets} shows that declaring the first exceedance
an onset is biased, and recommends change-point detection instead. We
retain the threshold-and-persistence form because the endpoint was
pre-registered, and note one difference in kind: our index is
computational depth rather than time, so the onset is a property of
what a fixed readout can recover rather than of when information comes
into existence.

\subsection{Uncertainty in Onset}

Uncertainty in
\[
\ell_q^*
\]
is estimated by resampling proteins while preserving the evaluation unit used for the corresponding familiarity regime.

For each bootstrap replicate, we recompute the full depth curve and its onset using the already-fixed thresholding rule.

We report:

\begin{itemize}[leftmargin=*]
    \item the estimated onset \(\ell_q^*\);
    \item its bootstrap confidence interval;
    \item the held-out false-positive rate of the null-calibration rule;
    \item sensitivity to the pre-specified value of \(\alpha\).
\end{itemize}

If the observed curve never satisfies the onset criterion, the onset is reported as unresolved rather than assigned to the final layer.

\subsection{Primary Depth-Generalization Contrasts}

The main onset contrast is
\[
\Delta_{\mathrm{onset}}
=
\ell_{NN}^*
-
\ell_{FF}^*.
\]

A positive value indicates that pocket information becomes reliably accessible later for novel-family/novel-pocket targets than for familiar-family/familiar-pocket targets. This reading requires care, because a weaker signal crosses a fixed threshold later even when it begins at the same stage, and the \(NN\) regime is the regime expected to carry the weaker signal. The confound therefore points in the same direction as the hypothesis. Section~\ref{sec:onset_validation} bounds its size using paired synthetic arms that differ only in signal strength, and we report that benchmark alongside any measured \(\Delta_{\mathrm{onset}}\).

The two intermediate contrasts
\[
\ell_{FN}^*-\ell_{FF}^*
\]
and
\[
\ell_{NF}^*-\ell_{FF}^*
\]
separate the contribution of pocket novelty from protein-family novelty.

\section{Results in Full}
\label{sec:results-full}

\subsection{The separation is not explained by evaluation geometry}

Per-protein AUPRC depends on quantities that carry no pocket
information: a large pocket in a short chain is easier to score well
on than a small pocket in a long one. Table~\ref{tab:robustness-design}
regresses per-protein AUPRC at each feature set's peak layer on
protein length, pocket size and positive rate, on the familiarity
regime, and on both.

Two things follow, and they pull in different directions. The
familiarity regime explains variance the geometric covariates do not:
adding the regime indicators raises \(R^2\) by 0.080 to 0.090. But the
covariates explain between 3.7 and 3.8 times more than the regime does
beyond them. Most of what separates one protein's score from
another's is the shape of its evaluation problem.

That second observation does not undermine the regime contrast,
because the contrast is between cells rather than between proteins.
The \(FN\) and \(NF\) cells are closely matched on every covariate --
median lengths 382 and 367, pocket sizes 24.0 and 25.5, positive rates
0.068 and 0.071 -- and their median AUPRC differs by a factor of 2.6,
0.230 against 0.601. Geometry cannot account for a gap between two
groups that share it.

\begin{table}[t]
\centering
\caption{Per-protein AUPRC against design factors and familiarity regime, at each feature set's peak layer. \emph{design} is an ordinary least squares fit on protein length, pocket size and positive rate; \emph{regime} on indicators for the four familiarity cells. The last column is the variance the familiarity cells explain beyond the design covariates. The interpretation is in the text.}
\label{tab:robustness-design}
\begin{tabular}{lrrrrrr}
\toprule
Features & Layer & $n$ & $R^2_{\mathrm{design}}$ & $R^2_{\mathrm{regime}}$ & $R^2_{\mathrm{both}}$ & $\Delta R^2_{\mathrm{regime}\mid\mathrm{design}}$ \\
\midrule
\texttt{s} & 18 & 534 & 0.321 & 0.106 & 0.400 & 0.080 \\
\texttt{z} & 25 & 534 & 0.365 & 0.121 & 0.452 & 0.087 \\
concat. & 27 & 534 & 0.369 & 0.123 & 0.459 & 0.090 \\
\bottomrule
\end{tabular}
\end{table}

\subsection{The ordering is robust to how pocket familiarity is
defined; the magnitude is not}

The primary definition of pocket familiarity is
\texttt{pocket\_qcov} against pre-cutoff systems, aggregated over a
system's ligands by minimum and thresholded at 80. PLINDER supplies
two further pocket-similarity metrics, and
Table~\ref{tab:robustness-altsp} regroups the same per-protein scores
under \texttt{sp\_pli} at threshold 70, which reassigns 159 of 534
test systems. The probe is untouched: it was trained on pooled
training systems and never saw a cell label, so only the evaluation
grouping differs.

The direction survives. In all six estimable comparisons the
familiar-pocket cell remains above the novel-pocket cell. The
magnitude does not: on the primary feature set the \(FF\)--\(FN\) gap
falls from 0.300 to 0.087 and the \(NF\)--\(NN\) gap from 0.124 to
0.089, with the same pattern in the other two feature sets.

We therefore report the ordering as a finding and the size of the gap
as dependent on how pocket familiarity is operationalised. The
narrower \texttt{qcov} criterion appears to select a smaller and more
extreme set of novel pockets; whether those are less familiar or
simply harder is not settled by these data.

\begin{table}[t]
\centering
\caption{Cell medians under the primary definition of pocket familiarity and under an alternative: \texttt{sp\_pli} at threshold 70, aggregated by min over a system's ligands, which reassigns 159 of 534 test systems (30\%). Only the evaluation grouping differs; the probe never saw a cell label. The direction of the familiar-versus-novel pocket contrast is preserved in all 6 estimable comparisons. Its magnitude is smaller under the alternative. Cells below 20 systems are not reported.}
\label{tab:robustness-altsp}
\begin{tabular}{ll rr rr}
\toprule
 & & \multicolumn{2}{c}{primary} & \multicolumn{2}{c}{alternative} \\
\cmidrule(lr){3-4}\cmidrule(lr){5-6}
Features & Cell & $n$ & median & $n$ & median \\
\midrule
\texttt{s} & FF & 152 & 0.530 & 114 & 0.435 \\
 & FN & 122 & 0.230 & 160 & 0.348 \\
 & NF & 136 & 0.601 & 75 & 0.609 \\
 & NN & 124 & 0.476 & 185 & 0.520 \\
\addlinespace
\texttt{z} & FF & 152 & 0.695 & 114 & 0.596 \\
 & FN & 122 & 0.303 & 160 & 0.429 \\
 & NF & 136 & 0.704 & 75 & 0.710 \\
 & NN & 124 & 0.629 & 185 & 0.660 \\
\addlinespace
concat. & FF & 152 & 0.705 & 114 & 0.554 \\
 & FN & 122 & 0.353 & 160 & 0.422 \\
 & NF & 136 & 0.724 & 75 & 0.734 \\
 & NN & 124 & 0.637 & 185 & 0.665 \\
\bottomrule
\end{tabular}
\end{table}

\subsection{What the hard regime contains}

The regimes differ in performance; they may also differ in the kinds
of binding site they contain. Before inspecting any result we fixed a
set of ligand-type flags and structural quantities and compared
\(FN\) against \(NF\) using the same family-cluster bootstrap as
above.

Although the two are similar on protein length, positive rate, median alignment depth and the system-level pocket measure used in the
design-factor analysis, that similarity does not extend to every
representation of pocket composition. The median number of pocket
residues per ligand is lower in \(FN\) than in \(NF\), with the
interval for the contrast excluding zero
(Table~\ref{tab:fn-composition}). Cofactor and covalent composition
also differ: \(FN\) holds cofactor ligands far less often and covalent
ligands far more often. The remaining pre-specified quantities do not
separate the two.

So the gap cannot be described as holding between groups matched on
pocket size without qualification. They are similar on the coarse
system-level measure and differ in per-ligand composition, and those
are different estimands. We treat these differences as candidate
confounders of the familiarity axis rather than explanations of the
performance gap: no multiplicity correction is applied, and a
difference in what the cells contain is not a mechanism.

One flag declared in advance is absent. The study release provides
three separate oligo-related columns where one was planned, and
substituting any of them after seeing the schema would turn a
pre-specified comparison into a selected one, so it is omitted rather
than redefined.

\begin{table}[t]
\centering
\caption{Composition of the four familiarity regimes using four ligand-type flags and three structural quantities fixed before inspection of the corrected 2026-09 results. The primary contrast is \(FN-NF\); intervals come from 2000 resamples of receptor-family clusters. The earlier single oligo flag is omitted because that variable does not exist in the study-release schema and was not replaced post hoc. FN and NF are similar on the system-level design-factor pocket measure, but the per-ligand pocket-residue quantity tested here is a different estimand. 3 of 6 non-degenerate comparisons have cluster-bootstrap intervals excluding zero. Ligand atoms with crystal contacts takes the same value in every regime and is reported but not counted among non-degenerate comparisons. No multiplicity correction is applied; these differences are candidate compositional confounders, not causal mechanisms.}
\label{tab:fn-composition}
\begin{tabular}{l rrrr r l}
\toprule
Quantity & \(FF\) & \(FN\) & \(NF\) & \(NN\) & \(FN-NF\) & 95\% interval \\
\midrule
ligands in the structure & 2.000 & 3.000 & 2.000 & 3.000 & +1.000 & [+0.000, +2.000] \\
ligand atoms with crystal contacts & 0.000 & 0.000 & 0.000 & 0.000 & +0.000 & [+0.000, +0.000] \\
pocket residues$^{\star}$ & 22.000 & 18.250 & 22.000 & 18.500 & -3.750 & [-7.000, -0.500] \\
cofactor fraction$^{\star}$ & 0.349 & 0.115 & 0.412 & 0.250 & -0.297 & [-0.405, -0.199] \\
ion fraction & 0.178 & 0.189 & 0.221 & 0.282 & -0.032 & [-0.131, +0.061] \\
fragment fraction & 0.132 & 0.205 & 0.169 & 0.210 & +0.036 & [-0.060, +0.129] \\
covalent fraction$^{\star}$ & 0.204 & 0.254 & 0.110 & 0.129 & +0.144 & [+0.055, +0.240] \\
\bottomrule
\end{tabular}
\end{table}

\subsection{Covalency does not account for the deficit}

The composition difference suggests an explanation: covalent ligands
attach by forming a bond and often sit where no cavity presents
itself, so a cell holding more of them might be hard because it holds
them. We tested that and it does not hold.

Covalent systems are associated with lower decodability. Pooled across
the four regimes and adjusting for regime, their mean AUPRC is lower
by \CovAdj\ \(\CovAdjCI\), close to the unadjusted difference of
\(\CovUnadj\). But the association is not common across regimes
(\(p=\CovP\)), so one coefficient describes none of them exactly
(Table~\ref{tab:covalent}).

The per-regime slopes close the account. They are most negative in
\CovMostNeg\ and least so in \CovLeastNeg: the regime whose
composition first suggested the explanation has the \CovFNPlace\ slope
of the four. Within \(FN\) itself the covalent and non-covalent halves
do not separate from \(NF\)'s by an amount that resolves
(\(\CovSplitEst\), \(\CovSplitCI\)).

Two cautions attach. AUPRC is bounded below and \CovLowestBase\ has
the lowest baseline of the four, so a negative association there has
least room to express itself and some of the heterogeneity may be
that. And these comparisons were decided after seeing the composition
result, so they are exploratory.

\begin{table}[t]
\centering
\caption{Covalent ligands and pocket decodability, by regime. Pooled across the four and adjusting for regime, covalent systems have mean AUPRC lower by \(0.221\) \([-0.279, -0.160]\), against \(-0.245\) unadjusted. The association is not common across regimes: the interaction terms reduce the residual sum of squares by more than a wild cluster bootstrap under the additive model produces (\(p=0.004\)), so one coefficient describes none of them exactly. \(FN\), whose composition first suggested the explanation, has the third most negative slope of the four. Slopes come from a model with a covalent-by-regime term; medians are descriptive. Exploratory: these comparisons were decided after the composition result.}
\label{tab:covalent}
\begin{tabular}{l rr rr r}
\toprule
Regime & \multicolumn{2}{c}{covalent} & \multicolumn{2}{c}{other} & slope \\
\cmidrule(lr){2-3}\cmidrule(lr){4-5}
 & \(n\) & median & \(n\) & median & \\
\midrule
\(FF\) & 31 & 0.113 & 121 & 0.579 & -0.333 \\
\(FN\) & 31 & 0.116 & 91 & 0.318 & -0.123 \\
\(NF\) & 15 & 0.650 & 121 & 0.596 & -0.079 \\
\(NN\) & 16 & 0.044 & 108 & 0.529 & -0.318 \\
\bottomrule
\end{tabular}
\end{table}

\subsection{Both axes as slopes, without the cells}

The four regimes rest on a threshold. Varying it leaves the ordering
intact, but a continuous model removes it rather than moving it, so we
fit per-protein AUPRC on both familiarity axes directly, with family
support on a log scale because the counts run across three orders of
magnitude.

Pocket familiarity predicts decodability \ConSpDir, \(\ConSp\)
\(\ConSpCI\) per standard deviation, so the cell result is not an
artefact of the coverage cutoff. Family support predicts it
\ConSfDir, \(\ConSf\) \(\ConSfCI\): within the range this test set
spans, more pre-cutoff relatives go with lower decodability. The
pocket axis is the larger of the two by a factor of about \ConRatio.
\ConInterNote.

The family slope needs its range stated. The stratified test set is
concentrated at low support. \ConZeroPct\ per cent of its
\ConNFit\ systems have no pre-cutoff relative, and its median count is
\ConMedTest, against \ConMedAll\ across all resolved targets, so
the association is estimated mostly between no support and modest
support, and says nothing about heavily represented families. It runs
opposite to what is reported for binding affinity \cite{mirage},
though on a different task and a different support definition, so we
draw no direct comparison.

Two further limits. \ConCeiling\ per cent of targets sit at or above
99 on the pocket measure, which compresses the top of its range. And
families accumulate structures partly because they are difficult, so
support may track difficulty rather than cause it.

\begin{table}[t]
\centering
\caption{Both familiarity axes fitted continuously, without the coverage threshold that defines the four regimes. Standardised main-effect coefficients describe a one-standard-deviation change in one axis at the mean of the other; the interaction is on the product of the two standardised axes. Raw coefficients use the corresponding centred natural units. Both axes are centred, so each main effect is the slope at the other axis's mean. Fitted on 534 test systems at layer 18, feature set \\texttt{s}, with intervals from 2000 resamples of receptor-family clusters. The interaction does not separate from zero, so each slope is read at the other axis's mean. The family slope is estimated over a test set concentrated at low support: 49 per cent of its systems have no pre-cutoff relative and its median count is 1, against 51 across all resolved targets.}
\label{tab:continuous}
\begin{tabular}{l rl rl}
\toprule
Term & \multicolumn{2}{c}{standardised} & \multicolumn{2}{c}{raw} \\
\cmidrule(lr){2-3}\cmidrule(lr){4-5}
\midrule
family support, \(\log_{10}(1+S_f)\)$^{\star}$ & -0.069 & [-0.091, -0.048] & -0.0824 & [-0.1084, -0.0573] \\
pocket familiarity, \(S_p\)$^{\star}$ & +0.110 & [+0.091, +0.129] & +0.0045 & [+0.0038, +0.0053] \\
interaction & +0.012 & [-0.007, +0.031] & +0.0006 & [-0.0003, +0.0015] \\
\bottomrule
\end{tabular}
\end{table}

\subsection{Alignment depth does not account for the gap}

The regime difference is present at the first block, and what enters
that block is not raw sequence: Boltz-1 runs its MSA module first, so
the input already carries processed evolutionary information. Weak
alignments are therefore the natural suspect, and the pair similar on
the design factors is where to look.

\(FN\) and \(NF\) have near-identical median alignment depths,
\MsaFnDepth\ and \MsaNfDepth\ sequences, and near-identical fractions
at the recording ceiling, \MsaFnCeil\ and \MsaNfCeil, while their
median AUPRC differs by a factor of \MsaAuprcRatio\
(Table~\ref{tab:msa-depth}). Across all \MsaN\ test proteins depth and
AUPRC are essentially uncorrelated: Spearman \MsaRhoAll\ overall and
\MsaRhoUncapped\ among the \MsaNUncapped\ systems below the ceiling,
with \MsaRhoOthers. And \MsaShallowRegime\ has the shallowest
alignments of the four at \MsaShallowDepth\ while scoring
\MsaShallowAuprc, above \(FN\).

Two limits bound this. Depths were recorded with a cap of
\MsaCeiling\ and \MsaNCeiling\ of \MsaN\ systems sit at it, so
differences among well-aligned systems are understated and the
uncapped correlation is the interpretable one. And depth is a weak
proxy for what matters: AlphaFold~3 is reported to depend on
phylogenetic diversity rather than sequence count
\cite{feldman2026alphainterp}, with a few divergent sequences
restoring accuracy where many near-identical ones do not. This excludes median alignment depth as the explanation, not
the alignments.

\begin{table}[h]
\centering
\caption{Alignment depth by familiarity regime, with pocket decodability for comparison. Depth is the minimum over a system's chains of its alignment's sequence count, recorded with a cap of 8,192; the ceiling column gives the fraction of each regime's systems at that cap, whose true depth is unknown. AUPRC is the median at the peak layer of the \texttt{s} feature set (layer 18). Of the pair similar on protein length, the system-level pocket measure and positive rate, \(NF\) has 1.04\(\times\) the median alignment depth of \(FN\), while their median AUPRC differs by a factor of 2.6.}
\label{tab:msa-depth}
\begin{tabular}{l r r r r}
\toprule
Regime & \(n\) & median depth & at ceiling & median AUPRC \\
\midrule
\(FF\) & 152 & 8,192 & 0.56 & 0.530 \\
\(FN\) & 122 & 7,692 & 0.44 & 0.230 \\
\(NF\) & 136 & 7,966 & 0.48 & 0.601 \\
\(NN\) & 124 & 5,156 & 0.34 & 0.476 \\
\bottomrule
\end{tabular}
\end{table}

\subsection{Depth curve summary}

Peak AUPRC, the depth at which each curve reaches 95\% of that peak,
\(\ell_{90}\), and the fold over each regime's own random-ranking
baseline, for all twelve curves. The body quotes the weakest and
strongest folds and the range of \(\ell_{90}\); this is the full
table behind them.

\begin{table}[H]
\centering
\caption{Depth curve summary. The near-peak layer is the first measured layer reaching 95\% of the maximum median AUPRC; $\ell_{90}$ is the first reaching 90\% of the gain available above layer 0. Fold values are relative to the random-ranking baseline.}
\label{tab:curve-summary}
\begin{tabular}{llrrrrrr}
\toprule
Features & Regime & $n$ & Layer 0 fold & Near-peak & $\ell_{90}$ & Peak AUPRC & Final fold \\
\midrule
\texttt{both} & FF & 152 & 5.63 & 17 & 17 & 0.706 & 8.25 \\
\texttt{both} & FN & 122 & 3.30 & 26 & 26 & 0.352 & 4.89 \\
\texttt{both} & NF & 136 & 7.54 & 16 & 19 & 0.732 & 9.95 \\
\texttt{both} & NN & 124 & 6.17 & 22 & 22 & 0.639 & 8.44 \\
\texttt{s} & FF & 152 & 3.08 & 15 & 15 & 0.540 & 5.29 \\
\texttt{s} & FN & 122 & 2.15 & 22 & 22 & 0.270 & 3.35 \\
\texttt{s} & NF & 136 & 4.18 & 14 & 14 & 0.609 & 7.69 \\
\texttt{s} & NN & 124 & 3.13 & 17 & 17 & 0.476 & 6.44 \\
\texttt{zpool} & FF & 152 & 5.86 & 14 & 19 & 0.707 & 8.14 \\
\texttt{zpool} & FN & 122 & 3.35 & 17 & 27 & 0.350 & 4.94 \\
\texttt{zpool} & NF & 136 & 7.42 & 18 & 21 & 0.742 & 9.77 \\
\texttt{zpool} & NN & 124 & 6.35 & 17 & 17 & 0.629 & 8.31 \\
\bottomrule
\end{tabular}
\end{table}

\subsection{Threshold sensitivity figure}

The same result as Table~\ref{tab:threshold-sensitivity}, drawn:
\(FN\) is lowest at every threshold tested, and the relative order of
the other three is not stable.

\begin{figure}[t]
\centering
\includegraphics[width=\textwidth]{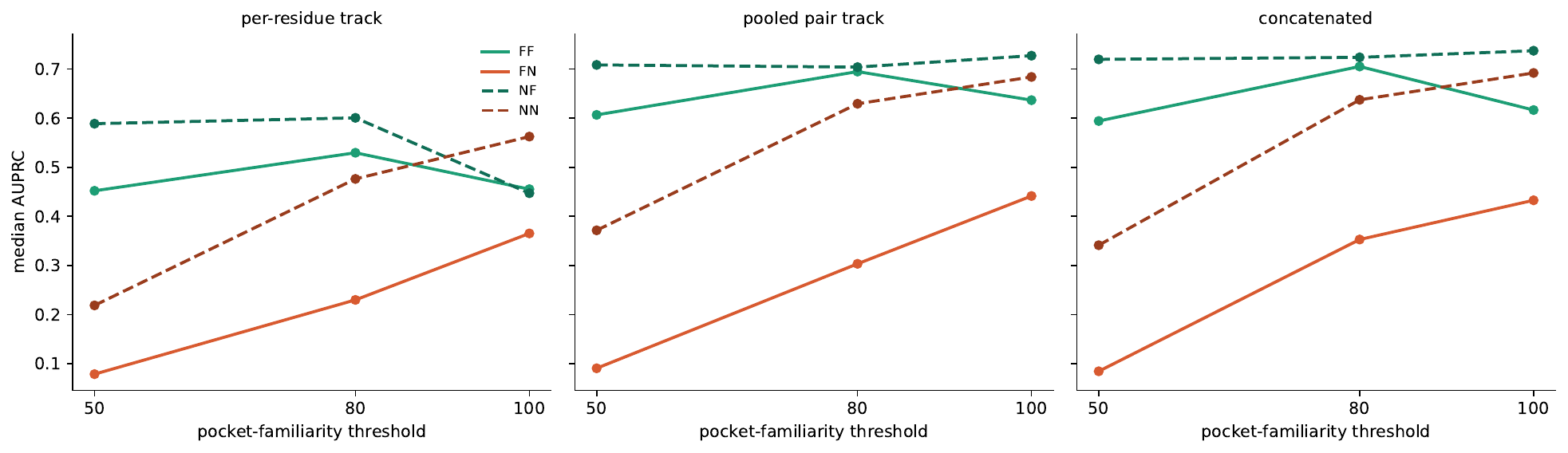}
\caption{Median AUPRC by regime against the pocket-familiarity
threshold, metric and aggregation held fixed. Hollow markers are cells
with fewer than 30 systems. \(FN\) is lowest throughout; the relative
order of the other three is not.}
\label{fig:threshold}
\end{figure}

\subsection{Bootstrap cells and contrasts}

Cell medians and all eighteen contrasts with their percentile
intervals, from 2000 resamples of receptor-family clusters. The
Holm-corrected versions of these contrasts are in
Table~\ref{tab:holm}; these are the uncorrected intervals the
correction was applied to.

\FloatBarrier
\begin{table}[H]
\centering
\caption{Median AUPRC per familiarity regime at each feature set's peak layer, with percentile intervals from 2000 bootstrap replicates. The resampling unit is the receptor-family cluster, not the protein: proteins within a family share sequence and fold, and resampling them individually would count near-duplicates as independent and give intervals that are too narrow.}
\label{tab:bootstrap-cells}
\begin{tabular}{llrrr}
\toprule
Features & Cell & $n$ & Median & 95\% interval \\
\midrule
\texttt{s} & FF & 152 & 0.530 & [0.410, 0.590] \\
 & FN & 122 & 0.230 & [0.139, 0.354] \\
 & NF & 136 & 0.601 & [0.542, 0.650] \\
 & NN & 124 & 0.476 & [0.374, 0.569] \\
\addlinespace
\texttt{z} & FF & 152 & 0.695 & [0.600, 0.729] \\
 & FN & 122 & 0.303 & [0.204, 0.414] \\
 & NF & 136 & 0.704 & [0.662, 0.761] \\
 & NN & 124 & 0.629 & [0.466, 0.710] \\
\addlinespace
concat. & FF & 152 & 0.705 & [0.581, 0.737] \\
 & FN & 122 & 0.353 & [0.177, 0.405] \\
 & NF & 136 & 0.724 & [0.671, 0.760] \\
 & NN & 124 & 0.637 & [0.479, 0.706] \\
\bottomrule
\end{tabular}
\end{table}

\begin{table}[t]
\centering
\caption{Bootstrap contrasts between familiarity regimes, each computed within the replicate rather than from two independently resampled estimates. Intervals exclude zero in every feature set for FF--FN, NF--FN, NN--FN. They contain zero in every feature set for FF--NN, NF--FF. The pre-registered primary contrast FF--NN is among the latter. 18 contrasts are reported with no multiplicity correction, so an interval excluding zero is an interval and not a test at a stated level. $\dagger$ marks the pre-registered primary contrast.}
\label{tab:bootstrap-contrasts}
\begin{tabular}{llrrc}
\toprule
Features & Contrast & Estimate & 95\% interval & Excludes 0 \\
\midrule
\texttt{s} & FF--FN & +0.300 & [+0.115, +0.403] & yes \\
 & NF--NN & +0.124 & [+0.018, +0.234] & yes \\
 & FF--NN$\dagger$ & +0.053 & [-0.096, +0.169] & no \\
 & NF--FF & +0.071 & [-0.015, +0.188] & no \\
 & NN--FN & +0.246 & [+0.070, +0.374] & yes \\
 & NF--FN & +0.371 & [+0.211, +0.470] & yes \\
\addlinespace
\texttt{z} & FF--FN & +0.392 & [+0.237, +0.504] & yes \\
 & NF--NN & +0.075 & [-0.017, +0.250] & no \\
 & FF--NN$\dagger$ & +0.066 & [-0.071, +0.231] & no \\
 & NF--FF & +0.009 & [-0.039, +0.120] & no \\
 & NN--FN & +0.326 & [+0.109, +0.479] & yes \\
 & NF--FN & +0.401 & [+0.281, +0.534] & yes \\
\addlinespace
concat. & FF--FN & +0.352 & [+0.215, +0.519] & yes \\
 & NF--NN & +0.086 & [+0.004, +0.258] & yes \\
 & FF--NN$\dagger$ & +0.068 & [-0.084, +0.234] & no \\
 & NF--FF & +0.019 & [-0.040, +0.146] & no \\
 & NN--FN & +0.285 & [+0.097, +0.483] & yes \\
 & NF--FN & +0.371 & [+0.298, +0.552] & yes \\
\bottomrule
\end{tabular}
\end{table}

\begin{figure}[t]
\centering
\includegraphics[width=0.78\textwidth]{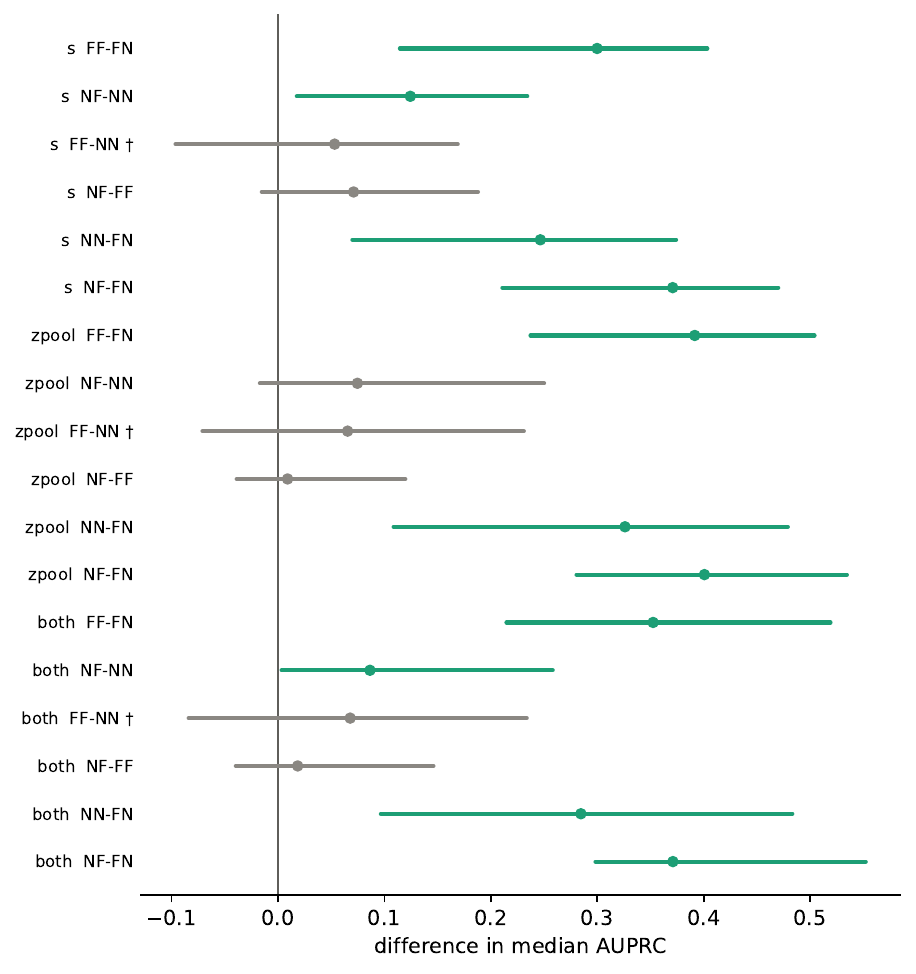}
\caption{All eighteen contrasts with their percentile intervals. Green
marks those excluding zero; $\dagger$ marks the pre-registered primary
contrast.}
\label{fig:bootstrap}
\end{figure}
\subsection{Onset and false-positive tables}

Table~\ref{tab:onset-observed} reports the onset under the
pre-registered rule for every regime and every feature set. It is
layer 0 throughout, with all 48 layers above the calibrated
threshold, so \(\Delta_{\mathrm{onset}}\) is zero by construction and
the endpoint the study was designed around cannot separate the
regimes. The table is included because a null endpoint is a result,
and because the uniformity of it — twelve curves, none of which
crosses its threshold later than the first block — is the thing that
makes the accumulation depth the informative statistic instead.

\begin{table}[H]
\centering
\caption{Onset of pocket decodability under the pre-registered rule: the first of 5 adjacent layers whose median AUPRC exceeds the null-calibrated threshold at its own layer. Every curve exceeds its threshold at every layer, so the onset is the first layer in all regimes and cannot distinguish them. The accumulation depth $\ell_{90}$, pre-declared as a secondary statistic for exactly this case, carries the depth comparison instead.}
\label{tab:onset-observed}
\begin{tabular}{lrrrr}
\toprule
Curve & Onset & First above $\tau$ & Layers above $\tau$ & $\ell_{90}$ \\
\midrule
\texttt{s/FF} & 0 & 0 & 48 of 48 & 15 \\
\texttt{s/FN} & 0 & 0 & 48 of 48 & 22 \\
\texttt{s/NF} & 0 & 0 & 48 of 48 & 14 \\
\texttt{s/NN} & 0 & 0 & 48 of 48 & 17 \\
\bottomrule
\end{tabular}
\end{table}

Table~\ref{tab:onset-fpr} reports how often each persistence
requirement fires on a null trajectory built from these
representations. Each trajectory is tested against a threshold
reconstructed from the other \(B-1\) replicates, so the trajectory
being judged contributes nothing to the threshold judging it. At the
pre-registered five stages the rate is 0.04 to 0.07 against a nominal
0.05; at three it is 0.12 to 0.17.

\begin{table}[H]
\centering
\caption{Empirical false-positive rate of the persistence rule on
this study's own null. $\star$ marks the pre-registered $p=5$.}
\label{tab:onset-fpr}
\begin{tabular}{lrrrrrrr}
\toprule
Curve & $p=1$ & $p=2$ & $p=3$ & $p=4$ & $p=5$ & $p=6$ & $p=7$ \\
\midrule
\texttt{s/FF} & 0.71 & 0.29 & 0.17 & 0.08 & 0.06$^{\star}$ & 0.04 & 0.02 \\
\texttt{s/FN} & 0.69 & 0.24 & 0.12 & 0.08 & 0.05$^{\star}$ & 0.05 & 0.04 \\
\texttt{s/NF} & 0.71 & 0.32 & 0.13 & 0.05 & 0.04$^{\star}$ & 0.03 & 0.02 \\
\texttt{s/NN} & 0.68 & 0.22 & 0.12 & 0.09 & 0.07$^{\star}$ & 0.04 & 0.02 \\
\bottomrule
\end{tabular}
\end{table}

The columns other than \(p=5\) are diagnostics rather than
alternative endpoints. The persistence value was fixed on synthetic
curves before any representation was extracted, and reporting the
rate at other values after the fact would invite reading the table as
a menu. It is included to show the shape of the failure: the error
rate rises steeply as the requirement weakens, which is what
correlation between neighbouring layers predicts and what a
per-layer test would not reveal.

\section{Design Specification}
\label{sec:design-full}

\subsection{Two-Axis Generalization Framework}

Let
\[
S_f
\]
denote protein-family familiarity and
\[
S_p
\]
denote pocket familiarity.

Each evaluation target is assigned to one of four regimes.

\FloatBarrier
\begin{table}[h]
\centering
\caption{Two-axis familiarity design.}
\begin{tabular}{lcc}
\toprule
 & \textbf{Familiar pocket} & \textbf{Novel pocket} \\
\midrule
\textbf{Familiar family} & FF & FN \\
\textbf{Novel family} & NF & NN \\
\bottomrule
\end{tabular}
\label{tab:quadrants}
\end{table}

The four regimes have distinct interpretations:

\begin{itemize}[leftmargin=*]
    \item \textbf{FF}: familiar protein family and familiar pocket geometry.
    \item \textbf{FN}: familiar family but structurally novel pocket.
    \item \textbf{NF}: novel family but a pocket resembling previously observed pockets.
    \item \textbf{NN}: novel family and novel pocket.
\end{itemize}

The pre-registered primary contrast is between \(FF\) and \(NN\). The
intermediate regimes are not decoration: \(FN\) and \(NF\) differ in
which axis is novel while closely sharing protein length, positive
rate and the system-level pocket measure, so a difference between them
is not attributable to those. The design therefore
admits a result in which the corner contrast is null while the
intermediate ones separate.

\subsection{Defining Protein-Family Familiarity}

Protein-family familiarity is defined from information available
before the evaluation target was deposited. Let \(S_f(i)\) denote the
historical support for target protein \(i\)'s family: the number of
pre-cutoff systems sharing a receptor-chain sequence cluster with it.
The construction follows \cite{mirage} in measuring support as a
count rather than a binary, and the clustering, the database snapshot
and the temporal cutoff were all fixed before any performance number
existed.

We retain both forms. \(S_f\) is thresholded into familiar and novel
for the four-cell evaluation, and kept as a count for the continuous
analysis, so the conclusions do not rest on the categorical form
alone. Appendix~\ref{sec:dataset-full} gives the construction, including the asymmetry between establishing familiarity and establishing
novelty.

\subsection{Defining Pocket Familiarity}
\label{sec:pocket-familiarity}

Let \(S_p(i)\) denote the structural similarity between target pocket
\(i\) and pockets available before the cutoff. It is based on a
pocket-level structural measure rather than protein-sequence identity,
since the two axes must measure different things for the design to
separate them.

As with family familiarity we retain both a continuous measure and a
thresholded designation for the \(2\times2\) evaluation. The
threshold was chosen from the sweep in Appendix~\ref{sec:dataset-full} on
cell balance, before performance numbers existed, and the ordering is
shown to survive moving it.

\subsection{Frozen Representation Setup}

Let
\[
H_\ell(x)
\]
denote a frozen representation of protein \(x\) at representation stage
\[
\ell = 0,\ldots,L.
\]

Depending on the trunk, the representation may include:

\begin{itemize}[leftmargin=*]
    \item per-residue single features \(s_\ell\),
    \item residue-pair features \(z_\ell\),
    \item or both.
\end{itemize}

The core study keeps the trunk frozen. This allows the experiment to ask what information is already accessible in the representation rather than conflating representation quality with end-to-end adaptation.

For any model family with only a final exported feature tensor, the final-stage analysis remains valid, but the onset analysis requires access to intermediate representation stages.

\section{Kill Criteria}

The study was to be reconsidered or narrowed if any of the following
occurred. None did. The fifth could not be evaluated, because the
readout comparison it refers to was not run.

\begin{enumerate}[leftmargin=*]

\item \(S_f\) and \(S_p\) are so strongly correlated that all four quadrants cannot be populated with meaningful sample sizes.

\item The \(NN\) regime is too small for stable pocket-level evaluation.

\item Performance shows no meaningful relationship with either familiarity axis.

\item The depth-onset analysis is unstable under bootstrap resampling or the null-calibration procedure fails to control held-out false positives.

\item Richer pair-processing heads improve performance uniformly across all regimes, providing no evidence of a novelty-specific readout bottleneck.

\end{enumerate}

\end{document}